\documentclass[journal]{IEEEtran}

\ifCLASSINFOpdf
\else
\fi
\usepackage{times}
\usepackage{soul}
\usepackage{url}
\usepackage{graphicx}
\usepackage{amsmath}
\usepackage{booktabs}
\usepackage{bbding}
\usepackage{amsfonts,amssymb}
\usepackage{multirow}
\usepackage{cite}
\usepackage{subfigure}
\begin{document}
%
\title{Utterance-to-Utterance Interactive Matching Network for Multi-Turn Response Selection in Retrieval-Based Chatbots}
%
%
%

\author{Jia-Chen~Gu, Zhen-Hua~Ling,~\IEEEmembership{Senior Member,~IEEE}, and Quan Liu
\thanks{This paper is the extended version of a short paper \cite{Gu:2019:IMN:3357384.3358140} that has been accepted by the ACM CIKM 2019 conference.}
\thanks{J.-C. Gu and Z.-H. Ling are with the National Engineering Laboratory for Speech and Language Information Processing, University of Science and Technology of China, Hefei 230027, China (e-mail: gujc@mail.ustc.edu.cn; zhling@ustc.edu.cn). Q. Liu is with the State Key Laboratory of Cognitive Intelligence, iFLYTEK Company, Ltd., Hefei 230088, China (e-mail: quanliu@iflytek.com).}
}

%
%

\markboth{PREPRINT MANUSCRIPT OF IEEE/ACM TRANSACTIONS ON AUDIO, SPEECH AND LANGUAGE PROCESSING}%
{Shell \MakeLowercase{\textit{et al.}}: Bare Demo of IEEEtran.cls for IEEE Journals}
%



\maketitle

\begin{abstract}
  This paper proposes an utterance-to-utterance interactive matching network (U2U-IMN) for multi-turn response selection in retrieval-based chatbots. Different from previous methods following context-to-response matching or utterance-to-response matching frameworks, this model treats both contexts and responses as sequences of utterances when calculating the matching degrees between them. For a context-response pair, the U2U-IMN model first encodes each utterance separately using recurrent and self-attention layers. Then, a global and bidirectional interaction between the context and the response is conducted using the attention mechanism to collect the matching information between them. The distances between context and response utterances are employed as a prior component when calculating the attention weights. Finally, sentence-level aggregation and context-response-level aggregation are executed in turn to obtain the feature vector for matching degree prediction. Experiments on four public datasets showed that our proposed method outperformed baseline methods on all metrics, achieving a new state-of-the-art performance and demonstrating compatibility across domains for multi-turn response selection.
\end{abstract}

\begin{IEEEkeywords}

dialogue, response selection, interactive matching network, utterance-to-utterance.
\end{IEEEkeywords}

%
\IEEEpeerreviewmaketitle

\section{Introduction}
%
%
%
%
  \IEEEPARstart{B}{uilding} a chatbot that can converse naturally with humans on open-domain topics is a challenging yet intriguing problem in artificial intelligence. Recently, human-computer conversation has attracted increasing attention due to its promising potential and commercial value \cite{DBLP:journals/sigkdd/ChenLYT17,DBLP:conf/aaai/YoungCCZBH18,DBLP:journals/corr/abs-1812-00686}. Existing approaches to building chatbots include generation-based methods \cite{DBLP:conf/acl/ShangLL15,DBLP:conf/aaai/SerbanSBCP16,DBLP:journals/www/ZhangZWZL19} and retrieval-based methods \cite{DBLP:conf/sigdial/LowePSP15,DBLP:journals/corr/KadlecSK15,DBLP:journals/dad/LowePSCLP17,DBLP:conf/acl/WuWXZL17,DBLP:conf/acl/WuLCZDYZL18,DBLP:conf/coling/ZhangLZZL18,gu-etal-2019-dually}. Response selection, which aims to select the best-matched response from a set of candidates given the context of a conversation, is the key technique for building retrieval-based chatbots.

  \begin{table}
  \small
  \caption{An example of a conversation in the Ubuntu V2 dataset whose response is composed of multiple utterances. ``\_eou\_" denotes end-of-utterance, and ``\_eot\_" denotes end-of-turn.}
  \label{tab1}
  \centering
  \begin{tabular}{l}
  \toprule
  \textbf{Conversation} \\
  \midrule
  \textbf{Speaker A:} How do I put myself in desktop in CUI? \_eou\_  \\
  \textbf{Speaker A:} I mean CLI. \_eou\_ \_eot\_     \\
  \textbf{Speaker B:} cd \~{}/ desktop. \_eou\_ \_eot\_ \\
  \textbf{Speaker A:} Is that the right code? cd / desktop? \_eou\_ \_eot\_   \\
  \midrule
  \textbf{Response Candidates} \\
  \midrule
  \textbf{Speaker B:} No. read it again. \_eou\_ Are you root? \_eou\_ That's\\
                      why new Ubuntu man's method will work for you. \_eou\_ \CheckmarkBold \\
  \textbf{Speaker B:} sebdc is talking nonsense. \_eou\_ You do not need \\
                      cpufreqd. \_eou\_ \XSolidBrush  \\
  \bottomrule
  \end{tabular}
  \end{table}

  In recent years, neural networks have been adopted to calculate the matching degrees between a context and its response candidates for response selection.
  Existing studies on neural network-based multi-turn response selection follow either context-to-response matching or utterance-to-response matching frameworks.
  The former adopts a coarse granularity for both contexts and responses that concatenates all utterances in a context or in a response into a single word sequence for matching degree calculation \cite{DBLP:conf/sigdial/LowePSP15,DBLP:journals/corr/KadlecSK15,DBLP:journals/dad/LowePSCLP17}.
  The latter adopts a fine granularity for contexts that separates a context into utterances but still concatenates all utterances in a response \cite{DBLP:conf/acl/WuWXZL17,DBLP:conf/acl/WuLCZDYZL18,DBLP:conf/coling/ZhangLZZL18}.
  However, both contexts and responses may contain multiple utterances in the response selection task, as illustrated in Table~\ref{tab1}.
  Both frameworks mentioned above neglect the relationships among the utterances in a response.

  Therefore, this paper proposes a neural network model named the utterance-to-utterance interactive matching network (U2U-IMN) for multi-turn response selection in retrieval-based chatbots.
  This model follows a new utterance-to-utterance (U2U) matching framework in order to deal with the situation in which both contexts and responses may contain multiple utterances.
  Different from the context-to-response matching and utterance-to-response matching frameworks, the U2U matching framework treats both contexts and responses as sequences of utterances when calculating the matching degrees between them.
  Therefore, the U2U-IMN model first encodes each utterance separately for a context-response pair.
  A previous study on natural language inference (NLI) \cite{DBLP:conf/acl/ChenZLWJI17} found that performing interactions between sentence pairs can provide useful matching information.
  Inspired by this, an attention-based interaction between the context and the response is conducted to collect the matching information between them.
  Here, the interaction is global (i.e., crossing utterance boundaries) and bidirectional (i.e., considering both context-to-response and response-to-context directions) in order to enrich the relevance representations of contexts and responses.
  The distances between context and response utterances are employed as a prior component when calculating the attention weights in order to distinguish the semantic contributions of different utterances in a context.
  Finally, sentence-level aggregation and context-response-level aggregation are executed in turn to obtain the feature vector for matching degree prediction.

  Our proposed methods were evaluated on two English datasets, the Ubuntu Dialogue Corpus V1 \cite{DBLP:conf/sigdial/LowePSP15} and Ubuntu Dialogue Corpus V2 \cite{DBLP:journals/dad/LowePSCLP17}, along with two Chinese datasets, the Douban Conversation Corpus \cite{DBLP:conf/acl/WuWXZL17} and E-commerce Dialogue Corpus \cite{DBLP:conf/coling/ZhangLZZL18}, which are all public datasets widely used in studies on multi-turn conversation.
  The results showed that our proposed method outperformed baseline methods on all metrics, achieved a new state-of-the-art performance, and demonstrated compatibility across domains for multi-turn response selection.

  In summary, the main contributions of this paper are twofold. First, this paper proposes a neural network model named U2U-IMN to deal with the situation in which both contexts and responses may contain multiple utterances.
  In this model, a matching module with attention-based global and bidirectional interactions is designed to collect the matching information between context and response utterances.
  Second, experimental results demonstrate that our proposed method achieves a new state-of-the-art performance on four public datasets for multi-turn response selection.

\section{Related Work}
  Chatbots aim to engage users in human-computer conversations in the open domain and are currently receiving increasing attention because they can target unstructured dialogue without a priori logical representation of the information exchanged during the conversation. Existing work on building chatbots includes generation-based methods \cite{DBLP:conf/acl/ShangLL15,DBLP:conf/aaai/SerbanSBCP16,DBLP:journals/www/ZhangZWZL19,DBLP:conf/acl/ZhuCZWL19,DBLP:conf/ijcai/SongZCWL19} and retrieval-based methods \cite{DBLP:conf/sigdial/LowePSP15,DBLP:journals/corr/KadlecSK15,DBLP:journals/dad/LowePSCLP17,DBLP:conf/acl/WuWXZL17,DBLP:conf/acl/WuLCZDYZL18,DBLP:conf/coling/ZhangLZZL18}. Generation-based models maximize the probability of generating a response given the previous dialogue. This approach enables the incorporation of rich context when mapping between consecutive dialogue turns. Retrieval-based chatbots have the advantage of generating informative and fluent responses because they select a proper response for the current conversation from a repository by means of response selection algorithms.

  Early studies on retrieval-based chatbots focused on single-turn conversation \cite{DBLP:conf/emnlp/WangLLC13,DBLP:journals/corr/JiLL14}.
  Recently, researchers have extended their attention to multi-turn conversation, which is more practical for real applications.
  A straightforward approach to multi-turn conversation is to match a response with the literal concatenation of context utterances \cite{DBLP:conf/sigdial/LowePSP15,DBLP:journals/corr/KadlecSK15,DBLP:journals/dad/LowePSCLP17}.
  Then, a multi-view model \cite{DBLP:conf/emnlp/ZhouDWZYTLY16}, including an utterance view and a word view, was studied.
  Wu et al. \cite{DBLP:conf/acl/WuWXZL17} proposed the sequential matching network (SMN), which first matched the response with each context utterance and then accumulated the matching information using a recurrent neural network (RNN). Zhang et al. \cite{DBLP:conf/coling/ZhangLZZL18} employed self-matching attention to route the vital information in each utterance based on SMN.
  The method of constructed representations at different granularities with stacked self-attention \cite{DBLP:conf/acl/WuLCZDYZL18} has also been presented.

  Our proposed U2U-IMN model has three main differences from the studies mentioned above.
  (1) U2U-IMN adopts a more fine-grained utterance-to-utterance (U2U) matching framework, while previous studies followed the framework of either context-to-response matching or utterance-to-response matching.
  (2) U2U-IMN derives the matching information between contexts and responses through global and bidirectional interactions, while
  the interactions used in previous studies were usually local and unidirectional \cite{DBLP:conf/acl/WuWXZL17}.
  (3) U2U-IMN employs the distances between context and response utterances as a prior component for calculating the attention weights in the interactive matching module.

\section{Utterance-to-Utterance \\ Interactive Matching Network}

  \subsection{Model Overview}
    Given a dialogue dataset $\mathcal{D}$, an example of the dataset can be represented as $(c,r,y)$. Specifically, $c = \{u^c_1,u^c_2,...,u^c_{n_c}\}$ represents a context with $\{u^c_m\}_{m=1}^{n_c}$ as its utterances and $n_c$ as its utterance number.
    Similarly, $r = \{u^r_1,u^r_2,...,u^r_{n_r}\}$ represents a response candidate with $\{u^r_n\}_{n=1}^{n_r}$ as its utterances and $n_r$ as its utterance number.
    Here, both the context and the response may be composed of multiple utterances, and the utterances in $c$ and $r$ are both chronologically ordered. $y \in \{0,1\}$ denotes a label. $y=1$ indicates that $r$ is a proper response for $c$; otherwise, $y=0$. Our goal is to learn a matching model $g(c,r)$ from $\mathcal{D}$. For any context-response pair $(c,r)$, $g(c,r)$ measures the matching degree between $c$ and $r$. We learn $g(c,r)$ by minimizing the sigmoid cross-entropy on $\mathcal{D}$. Let $\Theta$ denote the set of model parameters. Then, the objective function $\mathcal{L}(\mathcal{D}, \Theta)$ of learning can be formulated as
    \begin{equation}
    \begin{aligned}
    \mathcal{L}(\mathcal{D}, \Theta) =  - \sum_{(c,r,y)\in \mathcal{D}} [& y log(g(c,r))  \\
                                       & + (1-y)log(1-g(c,r)) ].
    \end{aligned}
    \end{equation}

    \begin{figure*}
    \centering
    \includegraphics[width=17cm]{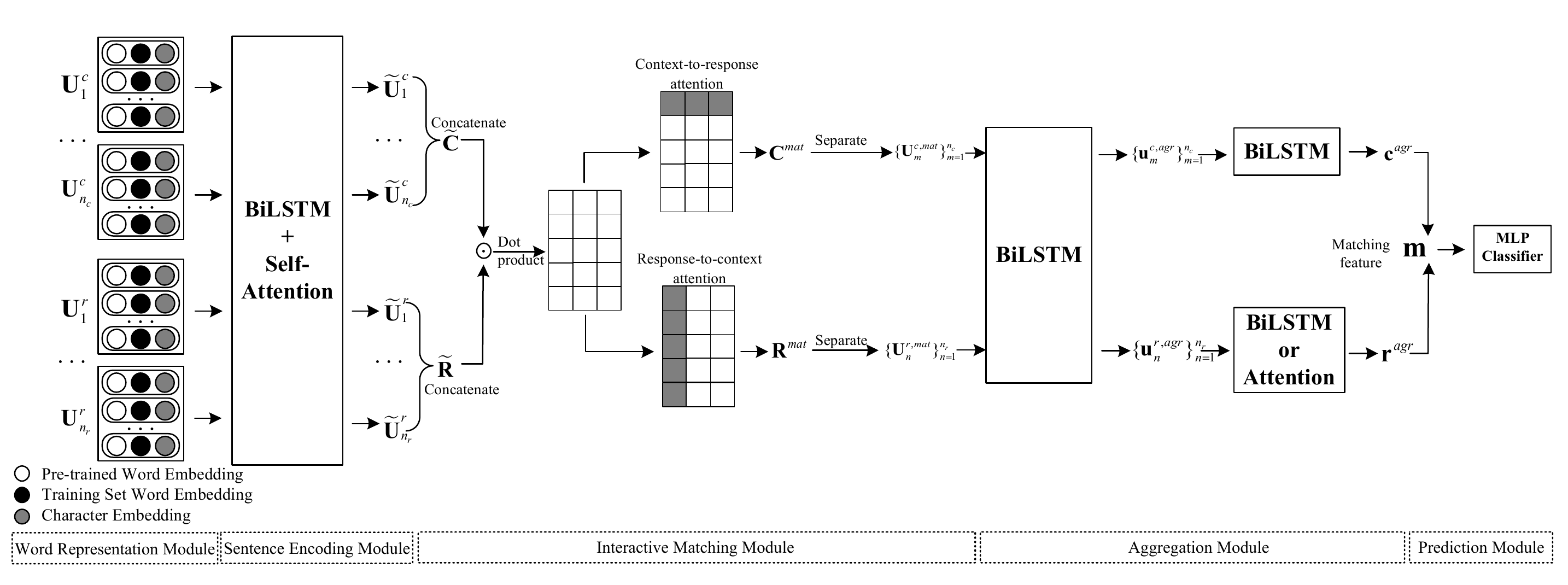}
    \caption{The overall architecture of our proposed U2U-IMN model.}
    \label{fig1}
    \end{figure*}

    The U2U-IMN model is designed to calculate the matching degree $g(c,r)$ for a context-response pair. It is composed of a word representation module, a sentence encoding module, an interactive matching module, an aggregation module and a prediction module, as shown in Fig.~\ref{fig1}. Details about each module are provided in the following subsections.

  \subsection{Word Representation Module}
    One challenge of word representation for dialogue is the large number of out-of-vocabulary (OOV) words.
    To address this issue, we combine the general pretrained word embeddings with those estimated on a task-specific training set \cite{DBLP:journals/corr/abs-1802-02614}.
    To further enhance the word embeddings, a convolutional neural network (CNN) is employed to model the morphology information at the character-level \cite{DBLP:conf/emnlp/LeeHLZ17}.

    Formally, the word embeddings of the \emph{m}-th utterance in a context and the \emph{n}-th utterance in a response candidate are denoted $\textbf{U}_m^c = \{\textbf{u}_{m,i}^c\}_{i=1}^{l_{u_m^c}}$ and $\textbf{U}_n^r = \{\textbf{u}_{n,j}^r\}_{j=1}^{l_{u_n^r}}$, respectively, where $l_{u_m^c}$ and $l_{u_n^r}$ are utterance lengths. Each $\textbf{u}_{m,i}^c$ or $\textbf{u}_{n,j}^r \in \mathbb{R}^d$ is an embedding vector of \emph{d} dimensions.

  \subsection{Sentence Encoding Module}
    First, each utterance in a context or in a response candidate is encoded by a bidirectional long short-term memory network (BiLSTM) \cite{DBLP:journals/neco/HochreiterS97}. We denote the calculations as follows:
    \begin{align}
    \bar{\textbf{u}}_{m,i}^c &= \textbf{BiLSTM}(\textbf{U}_m^c, i), i \in \{1, ..., l_{u_m^c}\},\\
    \bar{\textbf{u}}_{n,j}^r &= \textbf{BiLSTM}(\textbf{U}_n^r, j), j \in \{1, ..., l_{u_n^r}\}.
    \end{align}
    The parameters in these two BiLSTMs are shared in our implementation.

    To consider long-term dependency and highlight the semantic influences among adjacent words at the same time, a self-attention layer \cite{DBLP:conf/iclr/YuDLZ00L18} with a Gaussian prior \cite{guo2019gaussian} is employed to enhance the performance of BiLSTM-based sentence encoding.
    For a word in a context utterance, its representation after self-attention is calculated as
    \begin{equation}
    \tilde{\textbf{u}}_{m,i}^c = \sum_{j} \textbf{Softmax}(-|wd^2_{i,j} + b| + \bar{\textbf{u}}_{m,i}^{c\top} \cdot \bar{\textbf{u}}_{m,j}^c) \bar{\textbf{u}}_{m,j}^c,
    \label{equ7}
    \end{equation}
    where $d_{i,j}$ is the word-level distance between the $i$-th word and the $j$-th word, and $w$ and $b$ are scalar parameters estimated by model training.
    Similarly, for each word in a response utterance, we have
    \begin{equation}
    \tilde{\textbf{u}}_{n,j}^r = \sum_{i} \textbf{Softmax}(-|wd^2_{i,j} + b| + \bar{\textbf{u}}_{n,i}^{r\top} \cdot \bar{\textbf{u}}_{n,j}^r) \bar{\textbf{u}}_{n,i}^r.
    \end{equation}

    Finally, the outputs of the sentence encoding module are $\widetilde{\textbf{U}}_m^c=\{\tilde{\textbf{u}}_{m,i}^c\}_{i=1}^{l_{u_m^c}}, m \in \{1, ..., n_c\}$ for context utterances and $\widetilde{\textbf{U}}_n^r=\{\tilde{\textbf{u}}_{n,j}^r\}_{j=1}^{l_{u_n^r}}, n \in \{1, ..., n_r\}$ for response utterances.

  \subsection{Interactive Matching Module} \label{sec:imm}

    Interactions between the context and the response provide useful information for determining the matching degree between them.
    Unlike previous work \cite{DBLP:conf/acl/WuWXZL17,DBLP:conf/acl/WuLCZDYZL18,DBLP:conf/coling/ZhangLZZL18}, which matched the response to each utterance in the context separately, the U2U-IMN model matches the whole response with the whole context in a global and bidirectional way.
    Both the context and the response are treated as single word sequences, and attention weights are calculated between every word in the context and every word in the response.
    Then, the relevance representations are derived along both the context-to-response and response-to-context directions.
    This global and bidirectional strategy is expected to help neglect the irrelevant utterances and  enrich the relevance representations between the context and the response.
    Furthermore, considering that the context utterances adjacent to the response may contribute more in response selection than the distant ones, we propose to introduce an exponential prior based on the distance between context and response utterances when calculating the attention weights.

    First, the context representation $\widetilde{\textbf{C}} = [\tilde{\textbf{c}}_1,...,\tilde{\textbf{c}}_{l_c}]$ is formed by concatenating all context utterance representations $\{\widetilde{\textbf{U}}_m^c\}_{m=1}^{n_c}$, where $l_c = \sum_{m=1}^{n_c} l_{u_m^c}$ is the total number of words in the context.
    Similarly, we obtain $\widetilde{\textbf{R}} = [\tilde{\textbf{r}}_1,...,\tilde{\textbf{r}}_{l_r}]$ and $l_r = \sum_{n=1}^{n_r} l_{u_n^r}$ for the response.

    Then, an attention-based alignment is employed to collect relevance information between these two sequences by computing the attention weight between each pair of \{$\tilde{\textbf{c}}_{i}, \tilde{\textbf{r}}_j$\}
    as\footnote{Actually, the attention weights for context-to-response alignment and response-to-context alignment are different because of different normalization terms.
    Here, we use the same symbol $e_{ij}$, and the normalization term is not shown in Eq. (6) for simplification. }
    \begin{equation}
    e_{ij} \propto \phi(D_{i,j}) \cdot \exp(\tilde{\textbf{c}}_{i}^\top \cdot \tilde{\textbf{r}}_{j}),
    \label{equ1}
    \end{equation}
    where $D_{i,j}$ is the sentence-level distance between these two words, and
    $\phi(D) = e^{-W D + B}$ is an exponential prior with decay constant $W$ and initial value $e^B$. Here, $W$ and $B$ are model parameters that need to be estimated.

    Next, the attention weights $e_{ij}$ computed above are used to bidirectionally obtain the local relevance between a context and a response.
    For a word in the context, its context-to-response relevance representation carried by the response is composed using $e_{ij}$ as
    \begin{equation}
    \begin{aligned}
    \hat{\textbf{c}}_i = & \sum_{j} e_{ij} \tilde{\textbf{r}}_j \\
                       = & \sum_{j} \textbf{Softmax}(-W D_{i,j} + B + \tilde{\textbf{c}}_i^\top \cdot \tilde{\textbf{r}}_j) \tilde{\textbf{r}}_j,
    \label{equ8}
    \end{aligned}
    \end{equation}
    where the contents in $\{\tilde{\textbf{r}}_j\}_{j=1}^{l_r}$ relevant to $\tilde{\textbf{c}}_i$ are selected to form $\hat{\textbf{c}}_i$.
    The same calculation is also performed for each word in the response to form the response-to-context representation as
    \begin{equation}
    \begin{aligned}
    \hat{\textbf{r}}_j = & \sum_{i} e_{ij} \tilde{\textbf{c}}_i \\
                       = & \sum_{i} \textbf{Softmax}(-W D_{i,j} + B + \tilde{\textbf{c}}_i^\top \cdot \tilde{\textbf{r}}_j) \tilde{\textbf{c}}_i.
    \end{aligned}
    \end{equation}
    For the whole context and the whole response, we have $\widehat{\textbf{C}}=\{\hat{\textbf{c}}_i\}_{i=1}^{l_c}$ and  $\widehat{\textbf{R}}=\{\hat{\textbf{r}}_j\}_{j=1}^{l_r}$.
    Following a previous study on interactive matching for NLI \cite{DBLP:conf/acl/ChenZLWJI17}, we compute the differences and the element-wise products between \{$\widetilde{\textbf{C}}, \widehat{\textbf{C}}$\} and between \{$\widetilde{\textbf{R}}, \widehat{\textbf{R}}$\}. The differences and the element-wise products are then concatenated with the original vectors to obtain the enhanced representations as follows:
    \begin{align}
    \textbf{C}^{mat} &= [\widetilde{\textbf{C}}, \widehat{\textbf{C}}, \widetilde{\textbf{C}}-\widehat{\textbf{C}}, \widetilde{\textbf{C}} \odot \widehat{\textbf{C}}],\\
    \textbf{R}^{mat} &= [\widetilde{\textbf{R}}, \widehat{\textbf{R}}, \widetilde{\textbf{R}}-\widehat{\textbf{R}}, \widetilde{\textbf{R}} \odot \widehat{\textbf{R}}].
    \end{align}

    Thus far, the relevant information between the context and the response has been collected, which is further converted back to the matching matrices of separated utterances as
    \begin{align}
    \{\textbf{U}_m^{c,mat}\}_{m=1}^{n_c} = \textbf{Separate}( \textbf{C}^{mat} ),\\
    \{\textbf{U}_n^{r,mat}\}_{n=1}^{n_r} = \textbf{Separate}( \textbf{R}^{mat} ),
    \end{align}
    where the $\textbf{Separate}$ operation is conducted by segmenting the whole sequences of relevant information according to utterance length.

  \subsection{Aggregation Module} \label{sec5}
    The aggregation module converts the matching matrices of separated utterances into a final matching vector.
    Previous studies  \cite{DBLP:conf/acl/WuWXZL17,DBLP:conf/acl/WuLCZDYZL18,DBLP:conf/coling/ZhangLZZL18} adopted the utterance-to-response matching framework and only aggregated the matching matrices of utterances in a context.
    In contrast, the U2U-IMN model needs to conduct the aggregation operation for both the context and the response.

    First, the matching matrix $\textbf{U}_m^{c,mat}$ or $\textbf{U}_n^{r,mat}$ for each utterance is processed by a BiLSTM and aggregated by max pooling and last-hidden-state pooling operations. For the matching matrix $\textbf{U}_m^{c,mat}$ of each context utterance, the calculations are as follows:
    \begin{align}
    \textbf{u}_{m,i}^{c,utr} &= \textbf{BiLSTM}(\textbf{U}_m^{c,mat}, i), i \in \{1, ..., l_{u_m^c}\},\\
    \textbf{u}_{m}^{c,agr}   &= [\textbf{u}_{m,max}^{c,utr};\textbf{u}_{m,l_{u_m^c}}^{c,utr}], m \in \{1, ..., n_c\},
    \end{align}
    where $\textbf{u}_{m,max}^{c,utr}$ and $\textbf{u}_{m,l_{u_m^c}}^{c,utr}$ denote the results of max pooling and last-hidden-state pooling for the sequence of $\textbf{u}_{m,i}^{c,utr}$.
    The same calculations are also performed for the matching matrix $\textbf{U}_n^{r,mat}$ of each response utterance as follows:
    \begin{align}
    \textbf{u}_{n,j}^{r,utr} &= \textbf{BiLSTM}(\textbf{U}_n^{r,mat}, j), j \in \{1, ..., l_{u_n^r}\},\\
    \textbf{u}_{n}^{r,agr}   &= [\textbf{u}_{n,max}^{r,utr};\textbf{u}_{n,l_{u_n^r}}^{r,utr}], n \in \{1, ..., n_r\}.
    \end{align}
    The weights for these two BiLSTMs are shared in our implementation. Thus far, we have obtained two sets of utterance embeddings $\textbf{U}^{c,agr}=\{\textbf{u}_{m}^{c,agr}\}_{m=1}^{n_c}$ and $\textbf{U}^{r,agr}=\{\textbf{u}_{n}^{r,agr}\}_{n=1}^{n_r}$ for the context and the response, respectively. The next step is to convert them into aggregated context and response embeddings.

    The embedding vector of the context is derived in a way similar to the utterance-level aggregation method mentioned above. The utterance embeddings in $\textbf{U}^{c,agr}$ are sent into another BiLSTM following the chronological order of utterances in the context. Combined max pooling and last-hidden-state pooling operations are also performed to obtain the context embedding vector as
    \begin{align}
    \textbf{u}_{m}^{c,ctx} = \textbf{BiLSTM}&(\textbf{U}^{c,agr}, m), m \in \{1, ..., n_c\},\\
    \textbf{c}^{agr} &= [\textbf{u}_{max}^{c,ctx};\textbf{u}_{n_c}^{c,ctx}].
    \end{align}

    For the response, two aggregation strategies are designed in this paper.
    \subsubsection{RNN Aggregation}
      This is identical to the context aggregation in which the chronological relationships among utterances in the response are modelled. The operations can be written as
      \begin{align}
      \textbf{u}_{n}^{r,res} = \textbf{BiLSTM}&(\textbf{U}^{r,agr}, n), n \in \{1, ..., n_r\},\label{equ5}\\
      \textbf{r}^{agr} &= [\textbf{u}_{max}^{r,res};\textbf{u}_{n_r}^{r,res}].
      \end{align}

    \subsubsection{Attention Aggregation}
      Different from contexts that usually contain approximately ten utterances, a response is composed of much fewer utterances (see Fig.~\ref{fig3} in the next section for detailed statistics). We suppose that chronological relationships in short sequences are not as important as those in long sequences. Therefore, attention aggregation is designed to replace the RNN aggregation for deriving the response embedding vector. Mathematically, we have
      \begin{align}
      \label{equ6}
      \textbf{r}^{agr} = \sum_{n=1}^{n_r} w_n^{n_r}\textbf{u}_{n}^{r,agr},
      \end{align}
      where $w_n^{n_r}$ denotes softmax-normalized position-dependent utterance weights. During model training, the maximum number of utterances in a response $n_r^{max}$ is set manually. For each $n_r \in \{1,...,n_r^{max}\}$, a group of weights $\{w_1^{n_r},...,w_{n_r}^{n_r}\}$ is estimated with the constraint $\sum_{n=1}^{n_r} w_n^{n_r}=1$.

      The final matching feature vector is the concatenation of the context embedding vector and the response embedding vector:
      \begin{equation}
      \textbf{m} = [\textbf{c}^{agr};\textbf{r}^{agr}].
      \end{equation}

  \subsection{Prediction Module}
    The matching feature vector $\textbf{m}$ is then sent into a multi-layer perceptron (MLP) classifier. An MLP is a feedforward neural network estimated in a supervised manner using examples of features together with known labels. Here, the MLP is designed to predict whether a context-response pair matches appropriately according to the matching feature vector $\textbf{m}$. Finally, the MLP returns a score to denote the degree of matching.

\section{Experiments}

  \subsection{Datasets}

    \begin{table*}
    \small
    \caption{Statistics of the datasets for evaluating our proposed methods.}
    \label{tab2}
    \centering
    \begin{tabular}{c|ccc|ccc|ccc|ccc}
    \toprule
    \multirow{2}*{Dataset} & \multicolumn{3}{c|}{Ubuntu V1} & \multicolumn{3}{c|}{Ubuntu V2} & \multicolumn{3}{c|}{Douban}& \multicolumn{3}{c}{E-commerce} \\
                       & Train & Valid & Test   & Train & Valid & Test   & Train & Valid & Test  & Train & Valid & Test  \\
    \hline
    pairs               & 1M    & 0.5M  & 0.5M  & 1M    & 195k  & 189k  & 1M    & 50k   & 10k   & 1M    & 10k   & 10k   \\
    \hline
    positive:negative  & 1: 1  & 1: 9  & 1: 9  & 1: 1  & 1: 9  & 1: 9  & 1: 1  & 1: 1  & 1: 9 & 1: 1  & 1: 1  & 1: 9 \\
    \hline
    positive/context   & 1     & 1     & 1     & 1     & 1     & 1     & 1     & 1     & 1.18  & 1     & 1     & 1     \\
    \hline
    turns/context      & 8.44  & 2.66  & 2.65  & 6.29  & 5.86  & 6.03  & 6.69  & 6.75  & 5.95  & 5.51  & 5.48  & 5.64  \\
    \hline
    words/utterance    & 20.38 & 21.16 & 21.17 & 14.06 & 15.28 & 15.28 & 18.56 & 18.50 & 20.74 & 7.02  & 6.99  & 7.11  \\
    \bottomrule
    \end{tabular}
    \end{table*}

    Two English public multi-turn response selection datasets, the Ubuntu Dialogue Corpus V1 \cite{DBLP:conf/sigdial/LowePSP15} and Ubuntu Dialogue Corpus V2 \cite{DBLP:journals/dad/LowePSCLP17}, and two Chinese datasets, the Douban Conversation Corpus \cite{DBLP:conf/acl/WuWXZL17} and E-commerce Dialogue Corpus \cite{DBLP:conf/coling/ZhangLZZL18}, were adopted to evaluate our proposed methods.
    In our experiments, we followed the splits of training, validation, and test sets provided by the original authors of the four datasets.
    The Ubuntu Dialogue Corpus V1 and V2 contain multi-turn dialogues about Ubuntu system troubleshooting in English. Here, we adopted the version of the Ubuntu Dialogue Corpus V1 shared by Xu et al. \cite{DBLP:journals/corr/XuLWSW16}, in which numbers, paths and URLs were replaced by placeholders. Compared with the Ubuntu Dialogue Corpus V1, the training, validation and test dialogues in the V2 dataset were generated in different periods without overlap. Moreover, the V2 dataset discriminated between the end of an utterance (\_eou\_) and the end of a turn (\_eot\_). In both of the Ubuntu corpora, the positive responses are true responses from humans, and the negative responses are randomly sampled. The Douban Conversation Corpus was crawled from a Chinese social network on open-domain topics. It was constructed in a similar way to the Ubuntu corpus. The Douban Conversation Corpus collected responses via a small inverted-index system, and labels were manually annotated. The E-commerce Dialogue Corpus collected real-world conversations between customers and customer service staff from the largest e-commerce platform in China.
    Some statistics of these datasets are provided in Table~\ref{tab2}.

    \begin{figure}
    \centering
    \includegraphics[width=7cm]{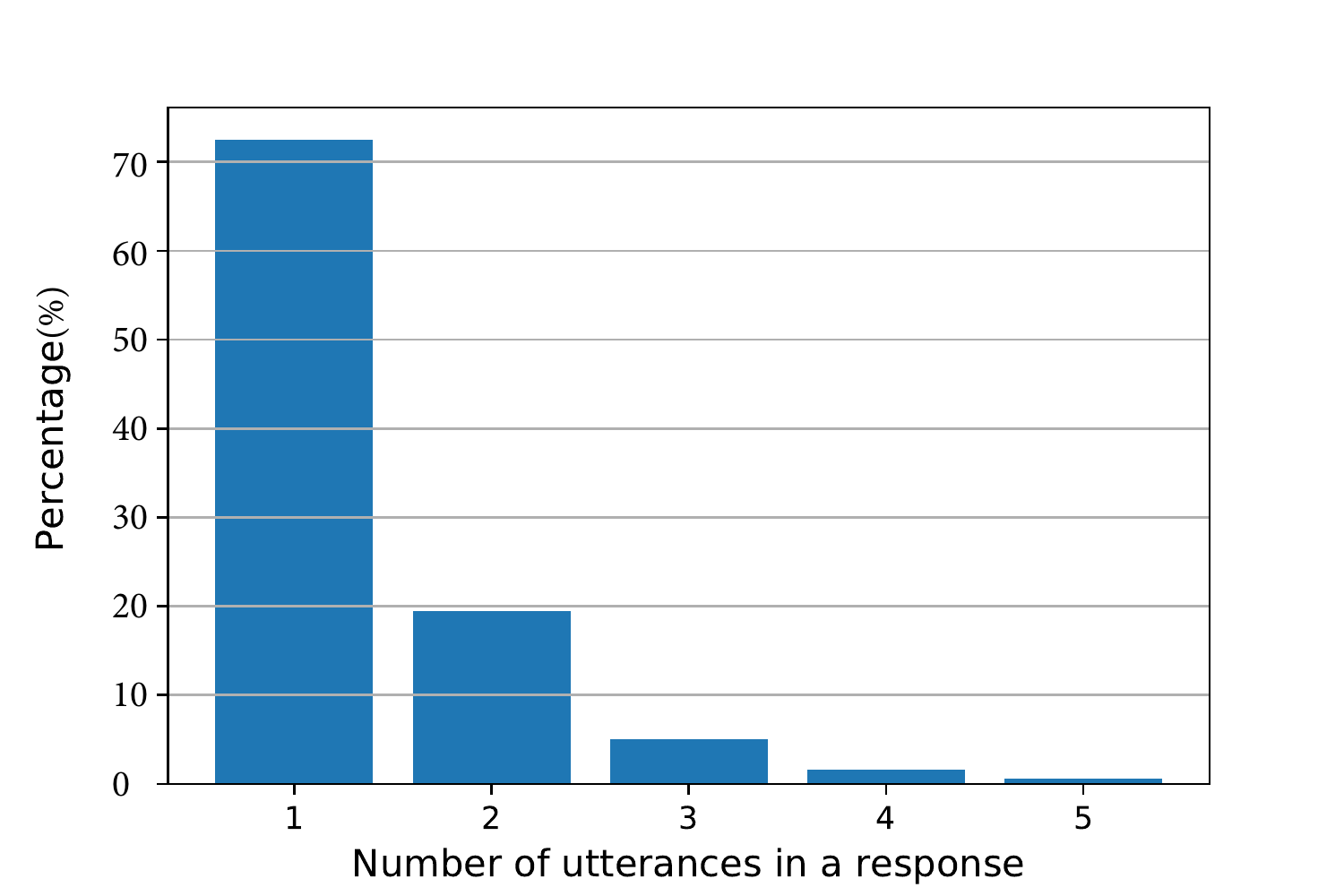}
    \caption{Distribution of responses in the Ubuntu Dialogue Corpus V2 across the number of utterances in a response.}
    \label{fig3}
    \end{figure}

    It is worth noting that the Ubuntu Dialogue Corpus V2 was the only dataset in our experiments that explicitly segmented utterances in responses. Specifically, approximately 30\% of the responses in this dataset consisted of multiple utterances, as shown in Fig.~\ref{fig3}, which made this dataset a very suitable one for evaluating our proposed U2U matching framework. The U2U-IMN model can also be applied to the other three datasets by considering a whole response as a single utterance.

  \subsection{Evaluation Metrics}
  The evaluation metrics  used in previous work \cite{DBLP:conf/sigdial/LowePSP15,DBLP:journals/dad/LowePSCLP17,DBLP:conf/acl/WuWXZL17,DBLP:conf/coling/ZhangLZZL18} were adopted in our experiments.
  Each model was tasked with selecting the $k$ best-matched responses from $n$ available candidates for the given conversation context $c$.
  We calculated the recall of the true positive replies among the $k$ selected responses, denoted $\textbf{R}_n@k$, as the main evaluation metric.
  The mean average precision (\textbf{MAP}) \cite{DBLP:books/aw/Baeza-YatesR99} was also adopted for reference since previous work did not list their results in terms of \textbf{MAP} on the Ubuntu V1, Ubuntu V2 and E-commerce datasets.
  In addition to $\textbf{R}_n@k$ and \textbf{MAP}, we also adopted the mean reciprocal rank (\textbf{MRR}) \cite{DBLP:conf/trec/Voorhees99} and precision-at-one ($\textbf{P}@1$) metrics for the Douban corpus, following the settings of previous work \cite{DBLP:conf/acl/WuWXZL17}.
  The reason was that the Douban Conversation Corpus was different from the other three datasets in that it included multiple correct candidates for a context in the test set, which may lead to low $\textbf{R}_{n}@k$.

  \subsection{Training Details}

    \begin{table*}[!hbt]
     \small
     \caption{Evaluation results of U2U-IMN and previous methods on the Ubuntu Dialogue Corpus V1 and V2.}
     \label{tab3}
     \centering
     \begin{tabular}{l|c|c|c|c|c|c|c|c|c|c}
      \toprule
                             & \multicolumn{5}{c|}{Ubuntu Corpus V1} & \multicolumn{5}{c}{Ubuntu Corpus V2} \\
      \hline
                             & \textbf{MAP} & $\textbf{R}_2@1$ & $\textbf{R}_{10}@1 $ & $\textbf{R}_{10}@2 $ & $\textbf{R}_{10}@5 $             & \textbf{MAP} & $\textbf{R}_2@1$ & $\textbf{R}_{10}@1 $ & $\textbf{R}_{10}@2 $ & $\textbf{R}_{10}@5 $\\
      \hline
       TF-IDF \cite{DBLP:conf/sigdial/LowePSP15,DBLP:journals/dad/LowePSCLP17}    &-& 0.659 & 0.410 & 0.545 & 0.708 &-& 0.749 & 0.488 & 0.587 & 0.763 \\
       RNN \cite{DBLP:conf/sigdial/LowePSP15,DBLP:journals/dad/LowePSCLP17}       &-& 0.768 & 0.403 & 0.547 & 0.819 &-& 0.777 & 0.379 & 0.561 & 0.836 \\
       LSTM \cite{DBLP:conf/sigdial/LowePSP15,DBLP:journals/dad/LowePSCLP17}      &-& 0.878 & 0.604 & 0.745 & 0.926 &-& 0.869 & 0.552 & 0.721 & 0.924 \\

       DL2R \cite{DBLP:conf/sigir/YanSW16}                &-& 0.899 & 0.626 & 0.783 & 0.944 &-& -     & -     & -     & -     \\
       Match-LSTM \cite{DBLP:conf/naacl/WangJ16}          &-& 0.904 & 0.653 & 0.799 & 0.944 &-& -     & -     & -     & -     \\
       MV-LSTM \cite{DBLP:conf/ijcai/WanLXGPC16}          &-& 0.906 & 0.653 & 0.804 & 0.946 &-& -     & -     & -     & -     \\
       Multi-View \cite{DBLP:conf/emnlp/ZhouDWZYTLY16}    &-& 0.908 & 0.662 & 0.801 & 0.951 &-& -     & -     & -     & -     \\
       RNN-CNN \cite{DBLP:journals/corr/BaudisS16}        &-& -     & -     & -     & -     &-& 0.911 & 0.672 & 0.809 & 0.956 \\
       CompAgg \cite{DBLP:journals/corr/WangJ16b}         &-& 0.884 & 0.631 & 0.753 & 0.927 &-& 0.895 & 0.641 & 0.776 & 0.937 \\
       BiMPM \cite{DBLP:conf/ijcai/WangHF17}              &-& 0.897 & 0.665 & 0.786 & 0.938 &-& 0.877 & 0.611 & 0.747 & 0.921 \\
       HRDE-LTC \cite{DBLP:conf/naacl/YoonSJ18}           &-& 0.916 & 0.684 & 0.822 & 0.960 &-& 0.915 & 0.652 & 0.815 & 0.966 \\
      \hline
       SMN \cite{DBLP:conf/acl/WuWXZL17}                  &-& 0.926 & 0.726 & 0.847 & 0.961 &-& -     & -     & -     & -     \\
       DUA \cite{DBLP:conf/coling/ZhangLZZL18}            &-& -     & 0.752 & 0.868 & 0.962 &-& -     & -     & -     & -     \\
       DAM \cite{DBLP:conf/acl/WuLCZDYZL18}               &-& 0.938 & 0.767 & 0.874 & 0.969 &-& -     & -     & -     & -     \\
      \hline
       U2U-IMN                                            & \textbf{0.866} & \textbf{0.945} & \textbf{0.790} & \textbf{0.886} & \textbf{0.973} & \textbf{0.852} & \textbf{0.943} & \textbf{0.762} & \textbf{0.877} & \textbf{0.975} \\
      \bottomrule
      \end{tabular}
    \end{table*}

    \begin{table*}[!hbt]
     \small
     \caption{Evaluation results of U2U-IMN and previous methods on the Douban Conversation Corpus and the E-commerce Corpus. All the results except ours are copied from \cite{DBLP:conf/acl/WuWXZL17,DBLP:conf/coling/ZhangLZZL18,DBLP:conf/acl/WuLCZDYZL18}.}
     \label{tab4}
     \centering
     \begin{tabular}{l|c|c|c|c|c|c|c|c|c|c}
      \toprule
                             & \multicolumn{6}{c|}{Douban Conversation Corpus} & \multicolumn{4}{c}{E-commerce Corpus} \\
      \hline
                             & \textbf{MAP} & \textbf{MRR} & $\textbf{P}@1$ & $\textbf{R}_{10}@1 $ & $\textbf{R}_{10}@2 $ & $\textbf{R}_{10}@5 $ & \textbf{MAP} & $\textbf{R}_{10}@1 $ & $\textbf{R}_{10}@2 $ & $\textbf{R}_{10}@5 $ \\
      \hline
       TF-IDF                & 0.331 & 0.359 & 0.180 & 0.096 & 0.172 & 0.405 &-& 0.159 & 0.256 & 0.477  \\
       RNN                   & 0.390 & 0.422 & 0.208 & 0.118 & 0.223 & 0.589 &-& 0.325 & 0.463 & 0.775  \\
       LSTM                  & 0.485 & 0.527 & 0.320 & 0.187 & 0.343 & 0.720 &-& 0.365 & 0.536 & 0.828  \\
       Multi-View            & 0.505 & 0.543 & 0.342 & 0.202 & 0.350 & 0.729 &-& 0.421 & 0.601 & 0.861  \\
       DL2R                  & 0.488 & 0.527 & 0.330 & 0.193 & 0.342 & 0.705 &-& 0.399 & 0.571 & 0.842  \\
       MV-LSTM               & 0.498 & 0.538 & 0.348 & 0.202 & 0.351 & 0.710 &-& 0.412 & 0.591 & 0.857  \\
       Match-LSTM            & 0.500 & 0.537 & 0.345 & 0.202 & 0.348 & 0.720 &-& 0.410 & 0.590 & 0.858  \\
      \hline
       SMN                   & 0.529 & 0.569 & 0.397 & 0.233 & 0.396 & 0.724 &-& 0.453 & 0.654 & 0.886  \\
       DUA                   & 0.551 & 0.599 & 0.421 & 0.243 & 0.421 & 0.780 &-& 0.501 & 0.700 & 0.921  \\
       DAM                   & 0.550 & 0.601 & 0.427 & 0.254 & 0.410 & 0.757 &-& -     & -     & -      \\
      \hline
      U2U-IMN                & \textbf{0.564} & \textbf{0.611} & \textbf{0.429} & \textbf{0.259} & \textbf{0.430} & \textbf{0.791}
                             & \textbf{0.759} & \textbf{0.616} & \textbf{0.806} & \textbf{0.966} \\
      \bottomrule
      \end{tabular}
    \end{table*}

    The Adam method \cite{DBLP:journals/corr/KingmaB14} was employed for optimization, with a batch size of 128.
    The initial learning rate was 0.001 and was exponentially decayed by 0.96 every 5000 steps.
    Dropout \cite{DBLP:journals/jmlr/SrivastavaHKSS14} with a rate of 0.2 was applied to the word embeddings and all hidden layers.

    The word representations for the English datasets were concatenations of the 300-dimensional GloVe embeddings \cite{DBLP:conf/emnlp/PenningtonSM14}, the 100-dimensional embeddings estimated on the training set using the Word2Vec algorithm \cite{DBLP:conf/nips/MikolovSCCD13} and the 150-dimensional character-level embeddings with window sizes of \{3, 4,  5\}, each consisting of 50 filters.
    The word embeddings for the Chinese datasets were concatenations of the 200-dimensional embeddings from previous work \cite{DBLP:conf/naacl/SongSLZ18} and the 200-dimensional embeddings estimated on the training set using the Word2Vec algorithm. Character-level embeddings were not employed for the two Chinese datasets due to the large number of Chinese characters.
    The word embeddings were not updated during training.

    All hidden states of LSTMs had 200 dimensions.
    The MLP of the prediction module had a hidden unit size of 256 with ReLU \cite{DBLP:conf/icml/NairH10} activation.
    The maximum word length, the maximum utterance length, the maximum number of utterances in a context, and the maximum number of utterances in a response were set as 18, 50, 10 and 3 respectively. We padded with zeros if the number of utterances in a context was less than 10 and the number of utterances in a response was less than 3.
    Otherwise, the last 10 utterances in the context or the last 3 utterances in the response were kept.
    The development set was used to select the best model for testing.

    All codes were implemented in the TensorFlow framework \cite{DBLP:conf/osdi/AbadiBCCDDDGIIK16} and have been published to help replicate our results\footnote{https://github.com/JasonForJoy/U2U-IMN}.

  \subsection{Experimental Results}

    Table~\ref{tab3} and Table~\ref{tab4} present the evaluation results of U2U-IMN and previous methods \footnote{In our previous conference paper, IMN employed an attentive hierarchical recurrent encoder (AHRE) as its sentence encoder, which aggregated multi-layer RNNs through attentive pooling.
    However, since the sentence encoder is not the key point of this paper, we replaced AHRE with a single-layer RNN for sentence encoding in U2U-IMN in order to simplify the model structure and focus on how to perform interactions between contexts and responses.}.

    All the results except ours are copied from the existing literature.
    For each dataset, all results listed in Table~\ref{tab3} or Table~\ref{tab4} are comparable with each other since they used the same training, validation and test data.
    Here, the U2U-IMN models adopted the attention aggregation strategy introduced in Section~\ref{sec5}.
    It can be observed from these two tables that U2U-IMN outperformed the other models on all metrics and datasets, which demonstrates its ability to select the correct response and its compatibility across domains (e.g., the domains of system troubleshooting, social networks and e-commerce covered by these datasets).

\section{Analysis}

 \begin{table}[hbt]
     \small
     \caption{Comparisons between U2R-IMN and U2U-IMN models on several subsets of the test set of the Ubuntu Dialogue Corpus V2. U2R-IMN denotes the model with concatenation of the utterances in a response. In each subset, the correct responses are composed of 1, 2, or 3 utterances.}
      \label{tab9}
     \centering
     \begin{tabular}{lccccc}
      \toprule
       Model            & Subset          & $\textbf{R}_2@1$ & $\textbf{R}_{10}@1 $ & $\textbf{R}_{10}@2 $ & $\textbf{R}_{10}@5 $\\
      \midrule
       U2R-IMN & 1 utt.                           & 0.936 & 0.733 & 0.863 & 0.974  \\
       U2U-IMN & 1 utt.                           & 0.937 & 0.737 & 0.863 & 0.972  \\
       \midrule
       U2R-IMN & 2 utt.                           & 0.952 & 0.823 & 0.904 & 0.979  \\
       U2U-IMN & 2 utt.                           & 0.956 & 0.831 & 0.911 & 0.984  \\
       \midrule
       U2R-IMN & 3 utt.                           & 0.965 & 0.873 & 0.923 & 0.982  \\
       U2U-IMN & 3 utt.                           & 0.976 & 0.904 & 0.955 & 0.994  \\
      \bottomrule
      \end{tabular}
    \end{table}

 \subsection{Effectiveness of U2U matching} \label{sec7}
    To further verify the effectiveness of our proposed U2U matching framework, we split the test set of the Ubuntu Dialogue Corpus V2 dataset according to the number of utterances in their correct responses.
    Then, the performances on these subsets of the U2U-IMN model were compared with those of the model (denoted U2R-IMN) that considered each response as a single utterance, as shown in Table~\ref{tab9}.
    As demonstrated, the U2U framework can help improve the performance by exploiting the relationships among the utterances in a response.
    We can see that the advantage of the U2U-IMN model over the U2R-IMN model became larger when the correct responses were composed of more utterances.
    This was consistent with the motivation of the U2U matching framework. Considering that only 30\% of responses in the Ubuntu Dialogue Corpus V2 dataset consisted of multiple utterances, a larger overall improvement may be achieved when applying our proposed U2U models to datasets containing more responses with multiple utterances.

  \subsection{Response aggregation strategies} \label{sec6}
    One key characteristic of the U2U matching framework is the response aggregation step that generates a single embedding vector based on the embedding vectors of response utterances.
    Table~\ref{tab5} shows the evaluation results of the two response aggregation strategies introduced in  Section~\ref{sec5}, where the \emph{RNN} suffix indicates the U2U-IMN model using the \emph{RNN} aggregation strategy instead of the attention aggregation.
    We can see than the U2U-IMN model with the default attention strategy for response aggregation achieved slightly better performance than that with RNN aggregation, which supported our assumption that the chronological relationships among utterances in short sequences may not be essential in the aggregation module.
    Some further analysis on these two aggregation strategies are given in the following.

    \begin{table}[t]
     \small
     \caption{Evaluation results of our proposed U2U matching framework on the test set of the Ubuntu Dialogue Corpus V2. The \emph{RNN} suffix of the U2U model denotes replacing the attention aggregation with the \emph{RNN} aggregation in the aggregation module.}
      \label{tab5}
     \centering
     \begin{tabular}{lcccc}
      \toprule
       Model                      & $\textbf{R}_2@1$ & $\textbf{R}_{10}@1 $ & $\textbf{R}_{10}@2 $ & $\textbf{R}_{10}@5 $\\
      \midrule
       U2U-IMN                                             & 0.943 & 0.762 & 0.877 & 0.975  \\
       U2U-IMN$_{\emph{RNN}}$                              & 0.942 & 0.758 & 0.875 & 0.974  \\
      \bottomrule
      \end{tabular}
    \end{table}

    \subsubsection{RNN Aggregation}

      \begin{figure}
      \centering
      \includegraphics[width=6cm]{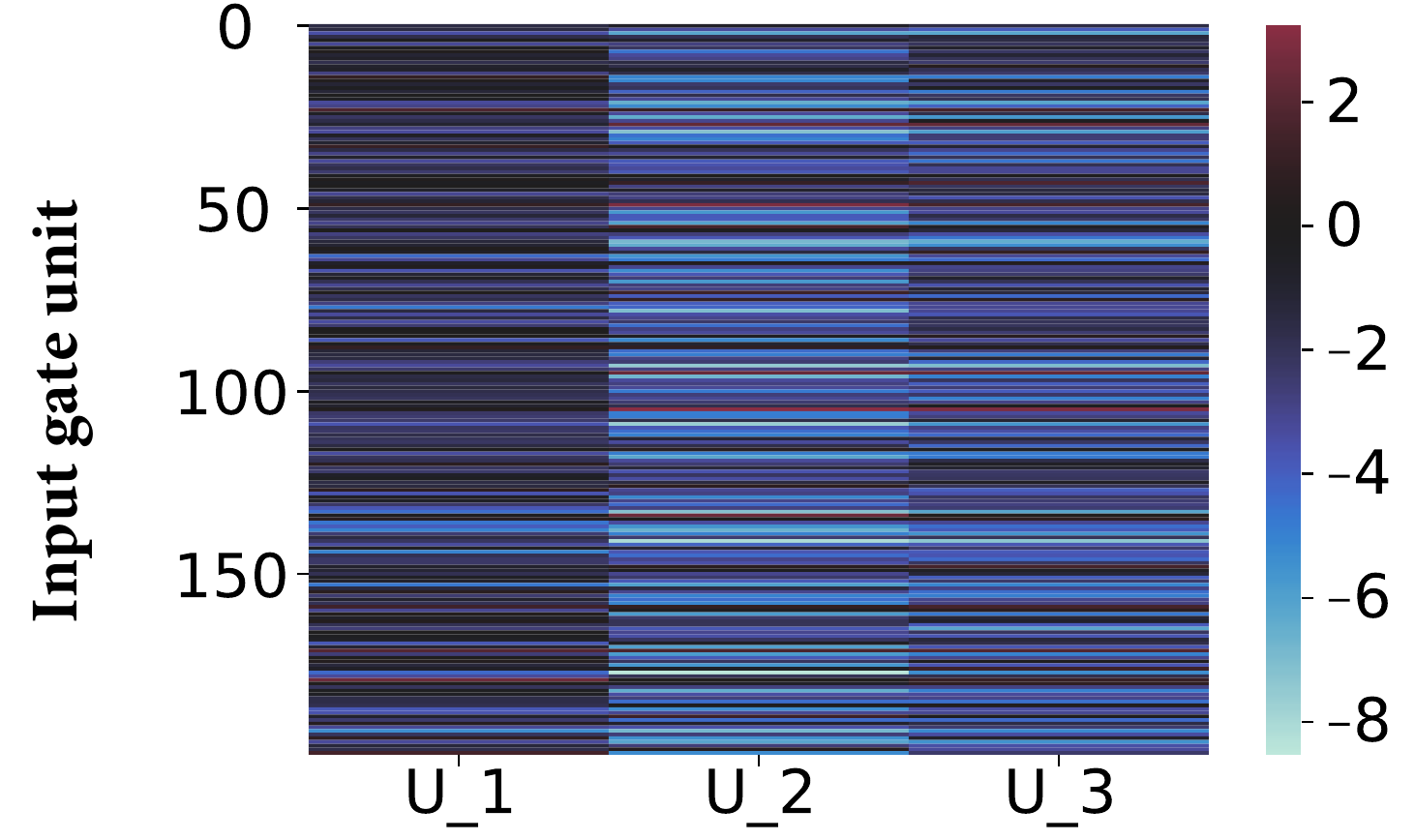}
      \caption{The input gate values of the LSTM in Eq.~(\ref{equ5}) of the U2U-IMN model for a response example in the test set of the Ubuntu Dialogue Corpus V2. The darker units correspond to larger values.}
      \label{fig7}
      \end{figure}

      To investigate how RNN aggregation identifies important utterances in a response, the input gate values of the LSTM in Eq.~(\ref{equ5}) for a response example were visualized, as shown in Fig.\ref{fig7}.
      The response was composed of three utterances,
      \emph{\{U$_1$: not as vboxnet0 though, windows names them local area connection \# 1,2,3 ... \_eou\_;
              U$_2$: exactly! \_eou\_;
              U$_3$: i don't know how to do it though :-lrb- \_eou\_ \}}, and
      \emph{U$_1$} was the most informative one. From Fig.\ref{fig7}, we can see that the input gates had larger values for \emph{U$_1$} than for the other two utterances. This means that more information from this utterance was preserved when aggregating the three utterances to form the embedding vector of the whole response.

    \subsubsection{Attention Aggregation}

      \begin{figure}
      \centering
      \includegraphics[width=6cm]{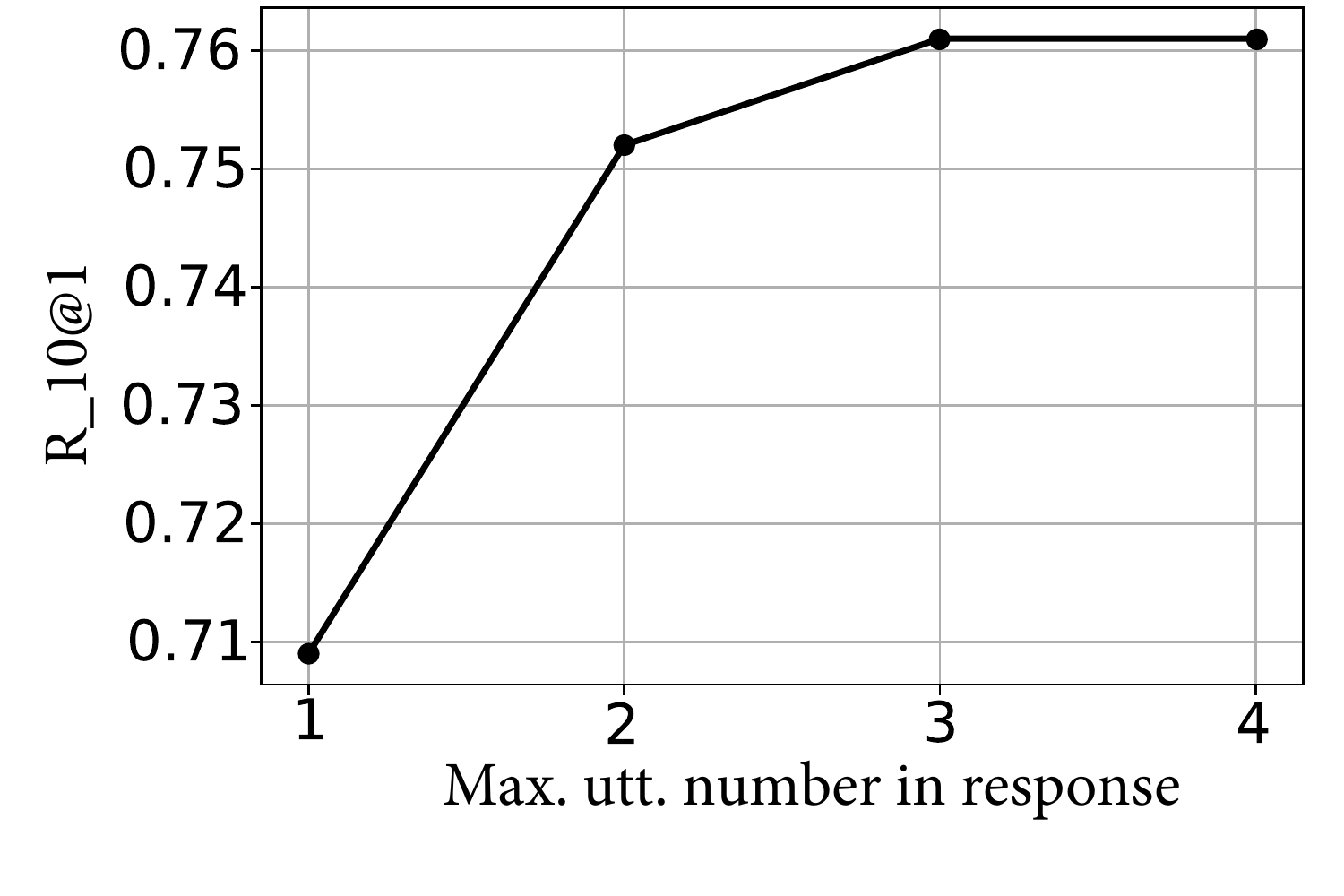}
      \caption{$\textbf{R}_{10}@1$ of U2U-IMN models with different $n_r^{max}$ tuned on the validation set of the Ubuntu Dialogue Corpus V2.}
      \label{fig8}
      \end{figure}

      \begin{table}[t]
      \small
      \caption{Attention weights $w_n^{n_r}$ of the U2U-IMN model with $n_r^{max}=3$ estimated on the training set of the Ubuntu Dialogue Corpus V2. }
      \label{tab8}
      \centering
      \begin{tabular}{cccc}
      \toprule
      $n_r$   & $w_1^{n_r}$ & $w_2^{n_r}$ & $w_3^{n_r}$  \\
      \midrule
      1   & 1.0     & -       & -      \\
      2   & 0.5986  & 0.4014  & -      \\
      3   & 0.4495  & 0.3014  & 0.2491 \\
      \bottomrule
      \end{tabular}
      \end{table}

      The maximum number of utterances in a response, i.e., $n_r^{max}$ in Section~\ref{sec5}, was tuned on the validation set, and the optimal one for the U2U-IMN model was $n_r^{max}=3$, as shown in Fig.~\ref{fig8}. The estimated attention weights $w_n^{n_r}$ of the U2U-IMN model with $n_r^{max}=3$  are shown in Table~\ref{tab8}. We can see that when $n_r>1$,  each utterance in the response contributed to forming the final response embeddings, and the first utterance contributed more than the last one.
      As we can see from the first row of Table~\ref{tab8} and Eq.~(\ref{equ6}), if there was only one utterance in a response, then the U2U-IMN model degenerated to follow the conventional utterance-to-response matching framework.

  \subsection{Bidirectional and global interactive matching}

    The bidirectional and global interactive matching between the context and the response in the U2U-IMN model is expected to help collect matching information and make matching decisions. Ablation tests and visualizations of attention weights were performed to demonstrate the effectiveness of both the bidirectional matching and the global matching.

    \subsubsection{Bidirectional Matching}

    \begin{table}[!hbt]
    \small
    \caption{Ablation tests of the context-to-response (C2R) and response-to-context (R2C) representations in the U2U-IMN model on the four datasets.}
    \centering
    \begin{tabular}{c|l|c|c|c}
    \toprule
     Dataset                    & \multicolumn{1}{c|}{Model} & $\textbf{R}_{10}@1 $ & $\textbf{R}_{10}@2 $ & $\textbf{R}_{10}@5 $  \\
    \hline
    \multirow{4}{*}{Ubuntu V1}  & U2U-IMN                & 0.790 & 0.886 & 0.973  \\
                                & \ - C2R                & 0.774 & 0.876 & 0.968  \\
                                & \ - R2C                & 0.780 & 0.880 & 0.971  \\
                                & \ - C2R\&R2C           & 0.650 & 0.806 & 0.954  \\
    \hline
    \multirow{4}{*}{Ubuntu V2}  & U2U-IMN                & 0.762 & 0.877 & 0.975  \\
                                & \ - C2R                & 0.738 & 0.866 & 0.972  \\
                                & \ - R2C                & 0.749 & 0.871 & 0.972  \\
                                & \ - C2R\&R2C           & 0.608 & 0.786 & 0.956  \\
    \hline
    \multirow{4}{*}{Douban}     & U2U-IMN                & 0.259 & 0.430 & 0.791  \\
                                & \ - C2R                & 0.251 & 0.424 & 0.784  \\
                                & \ - R2C                & 0.250 & 0.429 & 0.785  \\
                                & \ - C2R\&R2C           & 0.188 & 0.352 & 0.744  \\
    \hline
    \multirow{4}{*}{E-commerce} & U2U-IMN                & 0.616 & 0.806 & 0.966  \\
                                & \ - C2R                & 0.575 & 0.774 & 0.957  \\
                                & \ - R2C                & 0.567 & 0.766 & 0.961  \\
                                & \ - C2R\&R2C           & 0.538 & 0.751 & 0.938  \\
    \bottomrule
    \end{tabular}
    \label{tab7}
    \end{table}

     \begin{figure}[t]
    \centering
    \subfigure[Context-to-response]{
    \includegraphics[width=6cm]{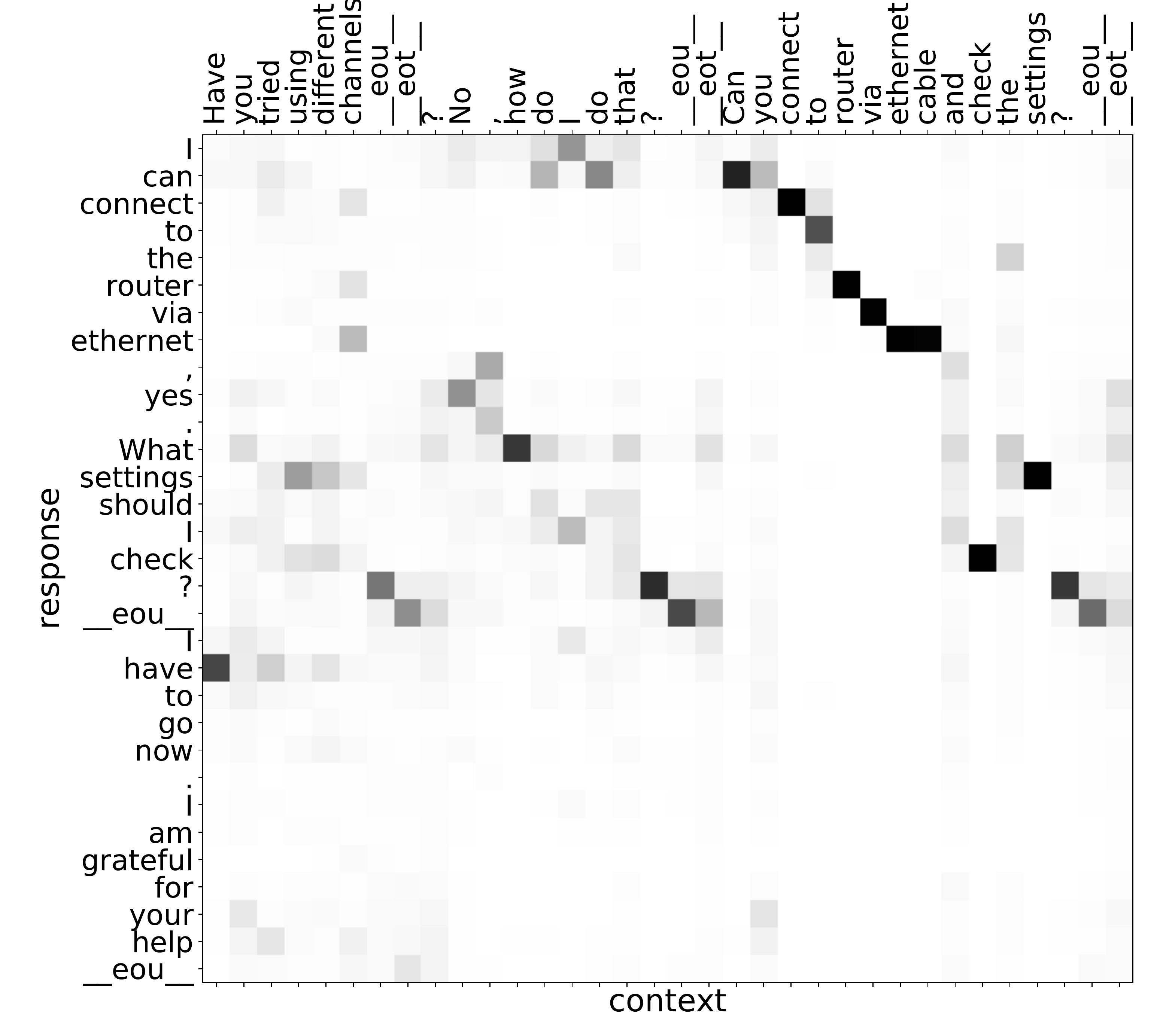}}
    \subfigure[Response-to-context]{
    \includegraphics[width=6cm]{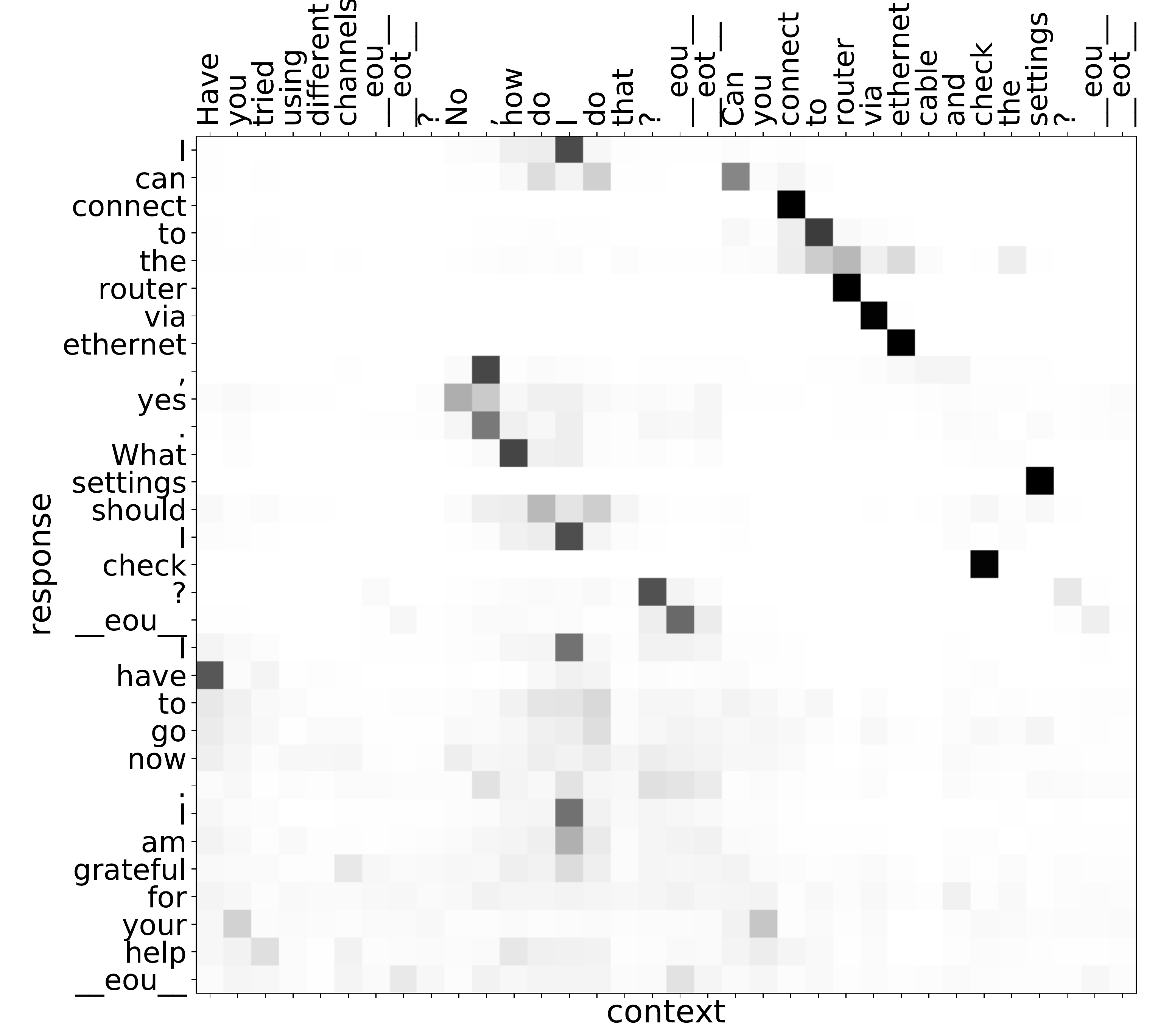}}
    \caption{Visualizations of the (a) context-to-response and (b) response-to-context attention weights in the interactive matching module for a test sample of the Ubuntu Dialogue Corpus V2. The darker units correspond to larger values.}
    \label{fig6}
    \end{figure}

    The bidirectional context-to-response and response-to-context representations in the U2U-IMN model were ablated. Specifically, when the context-to-response representation was ablated, the context representation given by the sentence encoding module $\{\widetilde{\textbf{U}}_m^c\}_{m=1}^{n_c}$ was sent to the aggregation module directly, and only the response representation $\{\widetilde{\textbf{U}}_n^r\}_{n=1}^{n_r}$ was enhanced by the interactive matching module to obtain $\{\textbf{U}_n^{r,mat}\}_{n=1}^{n_r}$ before aggregation.
    Similar operations were conducted to ablate the response-to-context representation.
    The results are shown in Table~\ref{tab7}. We can see that ablation of either the context-to-response or response-to-context representations resulted in a performance degradation, which indicates the effectiveness of the bidirectional matching between contexts and responses in the interactive matching module. A serious performance degradation can be observed when ablating the matching representations of both directions.

    A case study was further conducted by visualizing the bidirectional context-to-response and response-to-context attention weights for a test sample of the Ubuntu Dialogue Corpus V2.
    The context of the sample contained three utterances:
    \begin{itemize}
    \item \emph{Have you tried using different channels ? \_eou\_ \_eot\_}
    \item \emph{No, how do I do that ? \_eou\_ \_eot\_}
    \item \emph{Can you connect to router via ethernet cable and check the settings ? \_eou\_ \_eot\_}
    \end{itemize}
    The response was composed of two utterances:
    \begin{itemize}
    \item \emph{I can connect to the router via ethernet, yes  What settings should I check ? \_eou\_}
    \item \emph{I have to go now. I am grateful for your help \_eou\_}
    \end{itemize}
    The results are shown in Fig.~\ref{fig6}. We can see that some important words, such as ``\emph{connect}", ``\emph{router}" and ``\emph{ethernet}", in the context selected the relevant words in the response, and some unimportant words, such as ``\emph{grateful}", ``\emph{help}" and ``\emph{the}", in the response occupied small weights when forming the context-to-response representations. Identically, some important words in the response also selected the relevant words in the context, and some unimportant words in the context were also neglected when forming the response-to-context representations.

    \subsubsection{Global Matching}

    \begin{table}[t]
    \small
    \caption{Ablation tests of replacing the global context-response matching with local utterance-utterance matching on the four datasets.}
    \centering
    \begin{tabular}{c|c|c|c|c}
    \toprule
     Dataset                    & Model & $\textbf{R}_{10}@1 $ & $\textbf{R}_{10}@2 $ & $\textbf{R}_{10}@5 $  \\
    \hline
    \multirow{2}{*}{Ubuntu V1}  & U2U-IMN              & 0.790 & 0.886 & 0.973  \\
                                & - Global             & 0.786 & 0.885 & 0.972  \\
    \hline
    \multirow{2}{*}{Ubuntu V2}  & U2U-IMN              & 0.762 & 0.877 & 0.975  \\
                                & - Global             & 0.754 & 0.873 & 0.975  \\
    \hline
    \multirow{2}{*}{Douban}     & U2U-IMN              & 0.259 & 0.430 & 0.791  \\
                                & - Global             & 0.254 & 0.424 & 0.785  \\
    \hline
    \multirow{2}{*}{E-commerce} & U2U-IMN              & 0.616 & 0.806 & 0.966  \\
                                & - Global             & 0.586 & 0.792 & 0.961  \\
    \bottomrule
    \end{tabular}
    \label{tab10}
    \end{table}

    \begin{figure}[t]
    \centering
    \subfigure[Context to first response utterance]{
    \includegraphics[width=8cm]{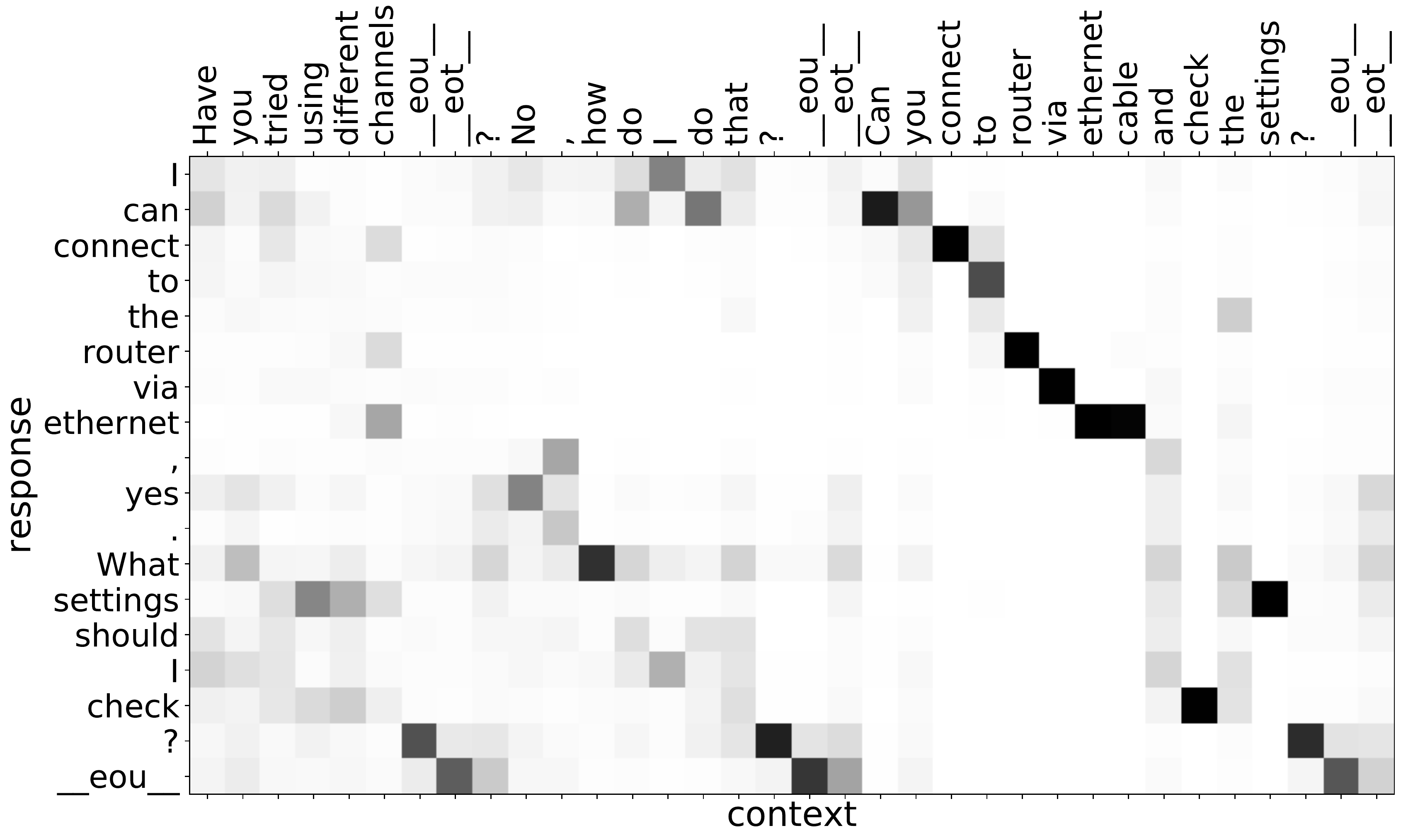}}
    \subfigure[Context to second response utterance]{
    \includegraphics[width=8cm]{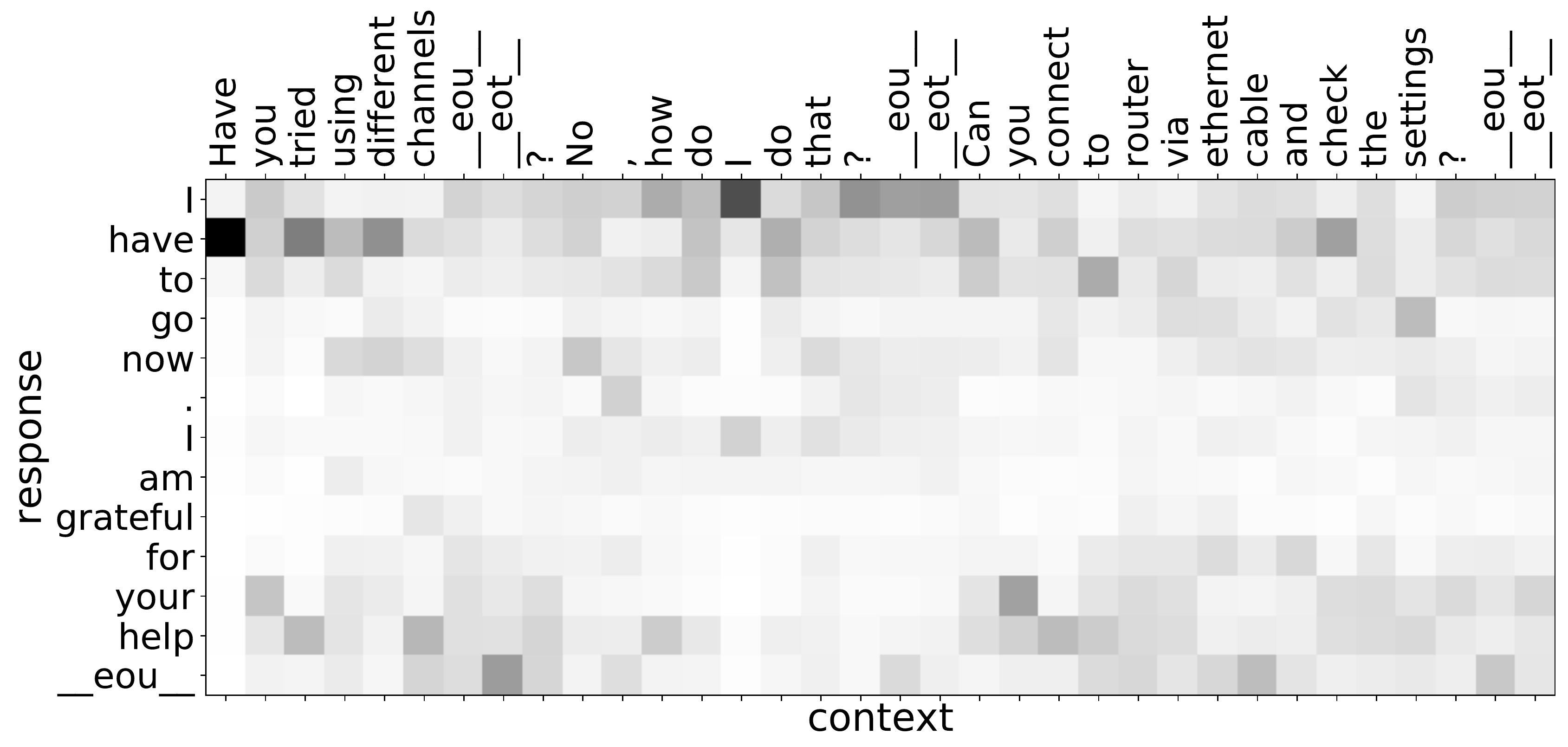}}
    \caption{Visualizations of attention weights between the context and each response utterance in the interactive matching module  for a test sample of the Ubuntu Dialogue Corpus V2. The darker units correspond to larger values.}
    \label{fig2}
    \end{figure}

    \begin{figure}[t]
    \centering
    \subfigure[Response to 1st context utterance]{
    \includegraphics[width=2.5cm]{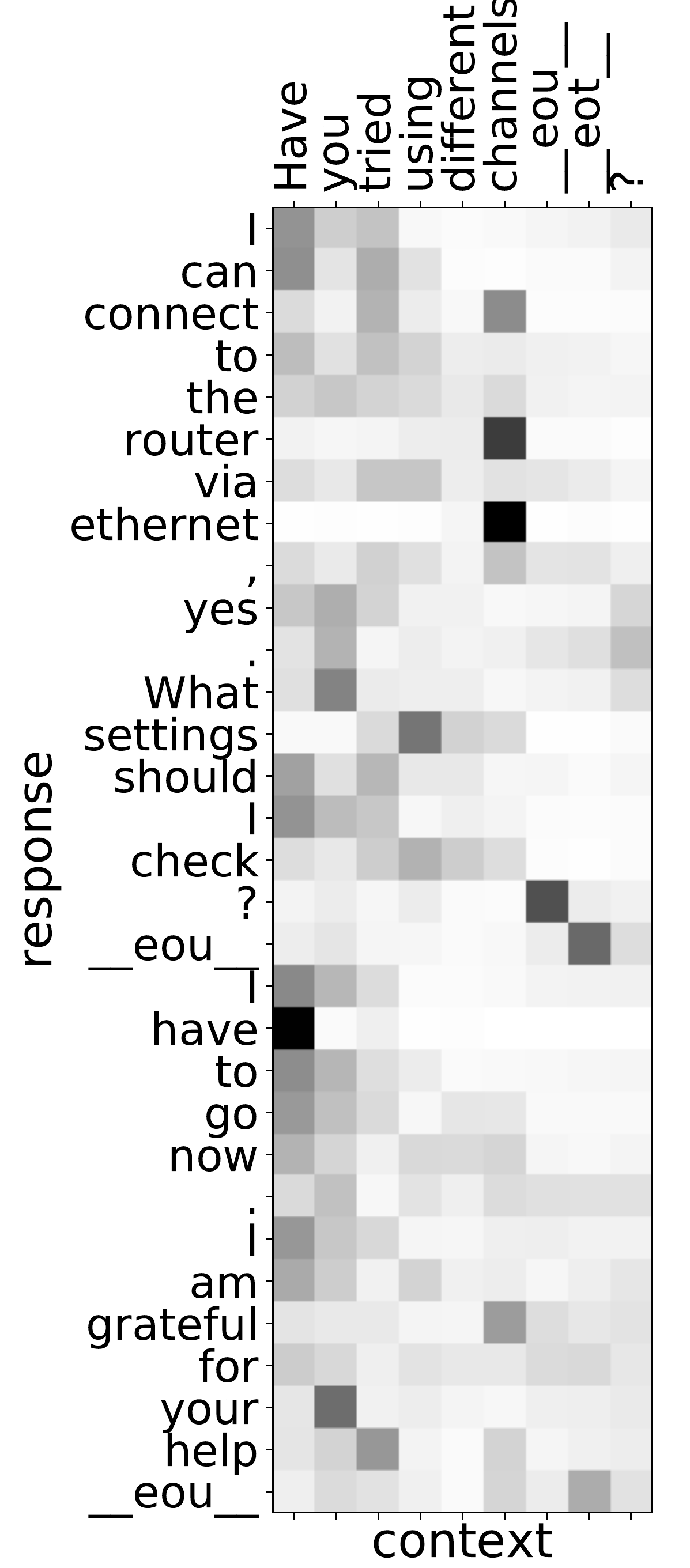}}
    \subfigure[Response to 2nd context utterance]{
    \includegraphics[width=2.5cm]{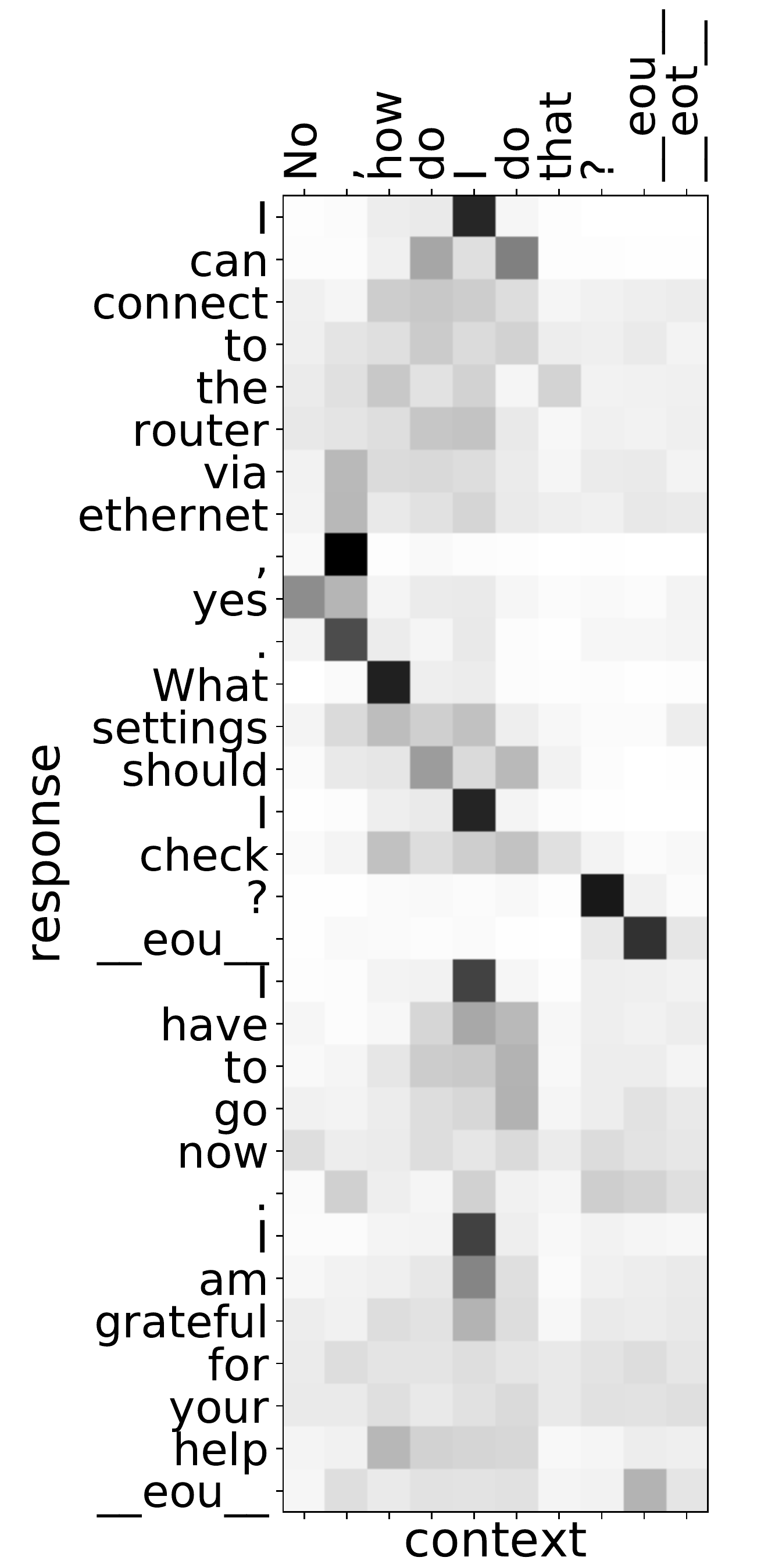}}
    \subfigure[Response to 3rd context utterance]{
    \includegraphics[width=3cm]{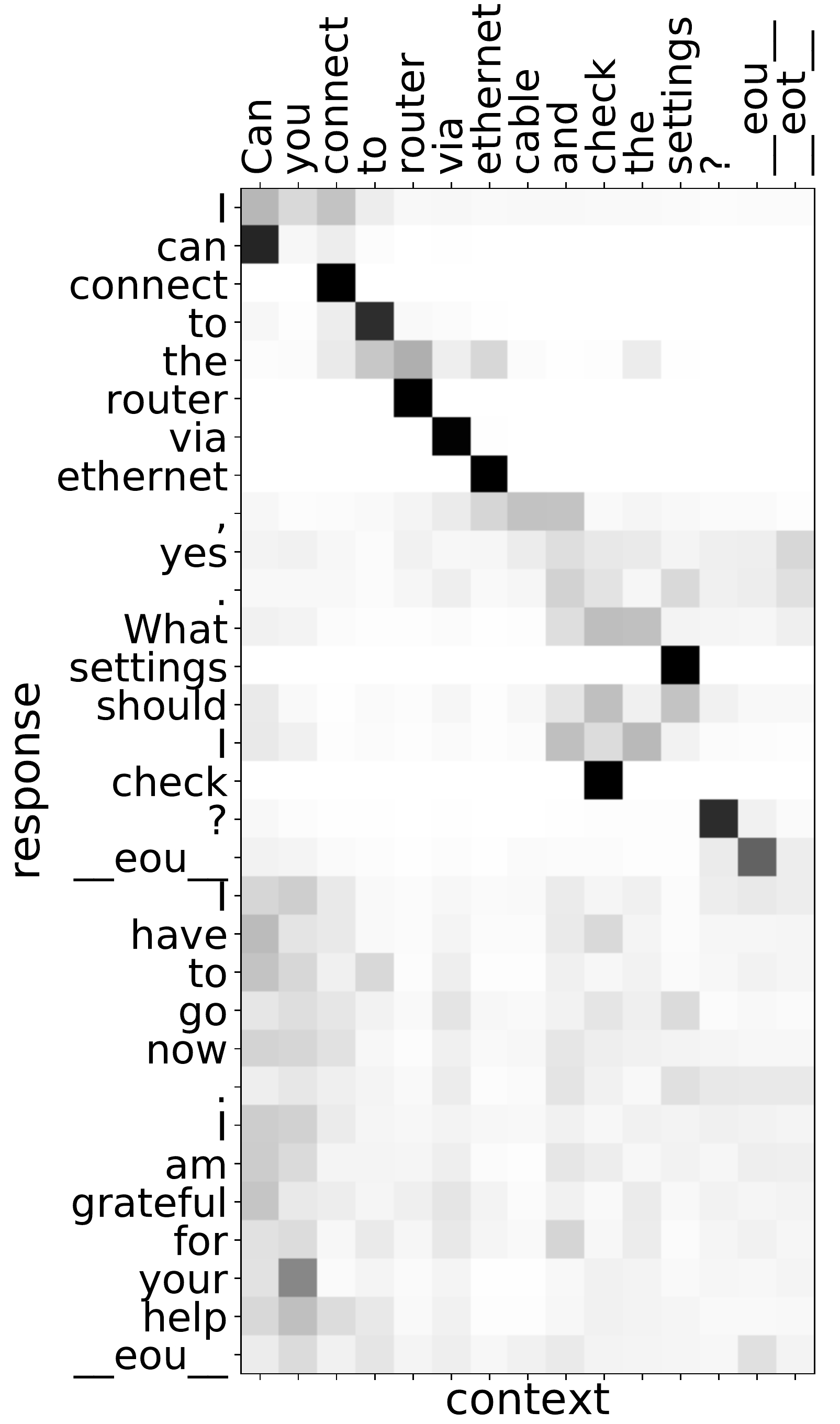}}
    \caption{Visualizations of attention weights between the response and each context utterance in the interactive matching module for a test sample of the Ubuntu Dialogue Corpus V2. The darker units correspond to larger values.}
    \label{fig5}
    \end{figure}

    To demonstrate the superiority of the global context-response matching used by the U2U-IMN model, an ablation test was conducted by  replacing it with local utterance-utterance matching.
    In the ablated model, the interactions introduced in Section \ref{sec:imm} were performed between each utterance in the context and each utterance in the response.
    Thus, we obtained a set of matching representations for each utterance in the context and each utterance in the response.
    Then, an additional pooling operation was performed over the set of representations to obtain the final matching representation for each utterance in the context and each utterance in the response.
    The pooling outputs were sent into the aggregation module for the following procedures.
    The results of the ablation test are shown in Table~\ref{tab10}, and the performance degradation demonstrated the superiority of our proposed global context-response matching to the local utterance-utterance matching in the interactive matching module.

    Furthermore, a case study was conducted by visualizing the context-to-utterance and response-to-utterance attention weights.
    The sample was the same as that used in Fig. \ref{fig6}.
    The results are shown in Fig.~\ref{fig2} and Fig.~\ref{fig5}, where the interactive matching was performed between the whole context and separated response utterances or between the whole response and separated context utterances.
    Comparing Fig.~\ref{fig6} (a) with Fig.~\ref{fig2} (b), we can see that the second response utterance  ``\emph{I have to go now. I am grateful for your help \_eou\_}" was less informative and occupies small weights in our proposed global context-response matching but occupies large weights in the context-to-utterance manner.
    The small weights of less informative utterances can help filter out irrelevant information in responses for deriving context representations.
    Similarly, comparing Fig.~\ref{fig6} (b) with Fig.~\ref{fig5} (a), we can find the same phenomenon for the first context utterance ``\emph{Have you tried using different channels \_eou\_ \_eot\_ ?}".
    These results verified the effectiveness of the global context-response interactive matching in our proposed U2U-IMN model.

  \subsection{Distance-based prior for interactive matching}

    \begin{table}[!hbt]
    \small
    \caption{Ablation tests of the distance-based prior for interactive matching on the four datasets.}
    \centering
    \begin{tabular}{c|c|c|c|c}
    \toprule
     Dataset                    & Model & $\textbf{R}_{10}@1 $ & $\textbf{R}_{10}@2 $ & $\textbf{R}_{10}@5 $  \\
    \hline
    \multirow{2}{*}{Ubuntu V1}  & U2U-IMN              & 0.790 & 0.886 & 0.973  \\
                                & - Prior              & 0.787 & 0.884 & 0.973  \\
    \hline
    \multirow{2}{*}{Ubuntu V2}  & U2U-IMN              & 0.762 & 0.877 & 0.975  \\
                                & - Prior              & 0.761 & 0.874 & 0.976  \\
    \hline
    \multirow{2}{*}{Douban}     & U2U-IMN              & 0.259 & 0.430 & 0.791  \\
                                & - Prior              & 0.251 & 0.431 & 0.782  \\
    \hline
    \multirow{2}{*}{E-commerce} & U2U-IMN              & 0.616 & 0.806 & 0.966  \\
                                & - Prior              & 0.600 & 0.795 & 0.968  \\
    \bottomrule
    \end{tabular}
    \label{tab6}
    \end{table}

    \begin{figure}[t]
    \centering
    \includegraphics[width=7cm]{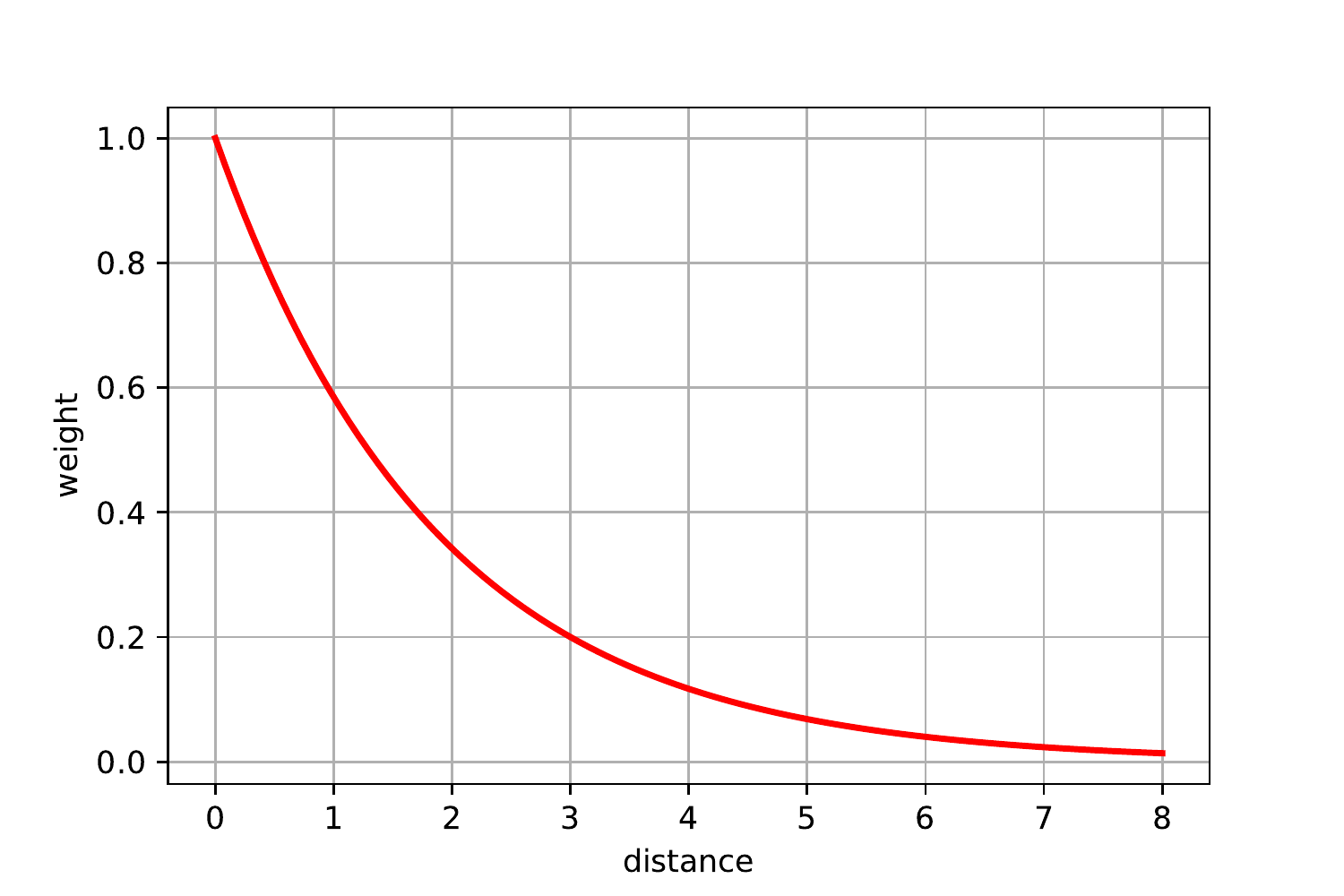}
    \caption{The estimated prior function $\Phi(D)=e^{-0.536D-0.00001}$ in Eq.~(\ref{equ8}) for the E-commerce Corpus.}
    \label{fig4}
    \end{figure}

    The exponential prior based on sentence-level distances in Eq. (6) of the interactive matching module was ablated, and the results on the test set of the four datasets are shown in Table~\ref{tab6}.
    We can see that the performance decreased on most metrics.
    Meanwhile, we can see that this distance-based prior provided larger improvements on the two Chinese datasets than on the two English datasets.
    The estimated prior function $\Phi(D)=e^{-0.536D-0.00001}$ in Eq.~(\ref{equ8}) for the E-commerce Corpus is drawn in Fig.~\ref{fig4}. We can see that larger weights were assigned to the utterances closer to the response.

\section{Conclusion}
  In this paper, we propose an utterance-to-utterance interactive matching network (U2U-IMN) for the multi-turn response selection task. Our proposed model first attempts to simultaneously explore the relationships among utterances in a context and those in a response. Then, U2U-IMN explores the matching information between contexts and responses through the global and bidirectional interactions between them. Meanwhile, distances are introduced into the interactions to distinguish the semantic contributions of utterances in a context according to their distances to the response. Experimental results show that our proposed model outperforms the baseline models on all metrics, achieving a new state-of-the-art performance and demonstrating compatibility across domains for multi-turn response selection in retrieval-based chatbots.
  Our future work includes
  (1) improving this proposed method to integrate more information, such as persona descriptions, for response selection,
  (2) applying the U2U framework to other matching scenes to further verify its effectiveness, and
  (3) employing pretrained models as effective resources for multi-turn response selection.

\section*{Acknowledgment}
  This work was partially funded by the National Key R\&D Program of China (Grant No. 2017YFB1002202), the National Nature Science Foundation of China (Grant No. 61871358, U1636201) and the Key Science and Technology Project of Anhui Province (Grant No.17030901005).


%

%
%
%

\ifCLASSOPTIONcaptionsoff
  \newpage
\fi

\bibliographystyle{IEEEtran}
\bibliography{IEEEtran}

\begin{IEEEbiography}[{\includegraphics[width=1in,height=1.25in,clip,keepaspectratio]{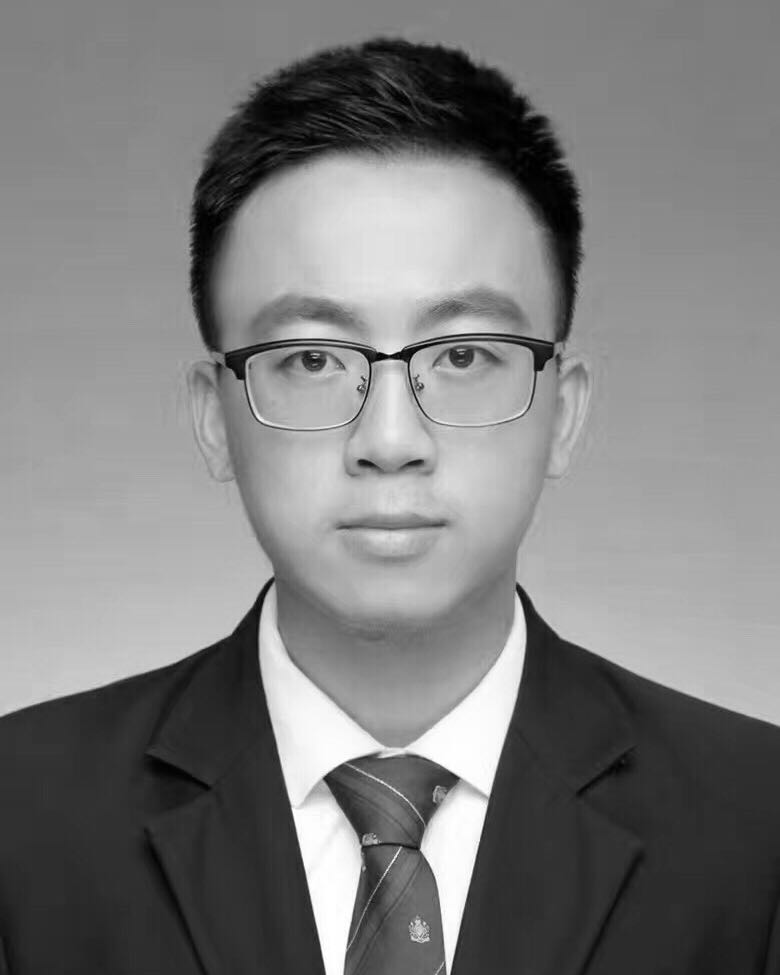}}]{Jia-Chen Gu}
  received a B.S. degree in communication engineering from Hohai University, Nanjing, China, in 2017. He is currently working towards a Ph.D. degree in signal and information processing at the University of Science and Technology of China, Hefei, China. His research interests include natural language understanding and deep learning.
\end{IEEEbiography}

\begin{IEEEbiography}[{\includegraphics[width=1in,height=1.25in,clip,keepaspectratio]{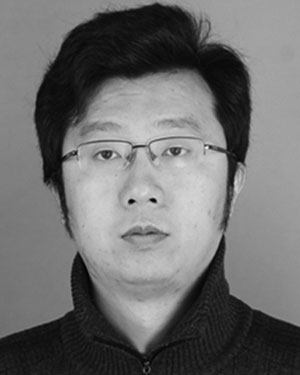}}]{Zhen-Hua Ling}
  (M' 10) received a B.E. degree in electronic information engineering and M.S. and Ph.D. degrees in signal and information processing from the University of Science and Technology of China, Hefei, China, in 2002, 2005, and 2008. From October 2007 to March 2008, he was a Marie Curie Fellow with the Centre for Speech Technology Research, University of Edinburgh, Edinburgh, U.K. From July 2008 to February 2011, he was a joint Postdoctoral Researcher with the University of Science and Technology of China, and iFLYTEK Company, Ltd., Hefei, China. He is currently an Associate Professor with the University of Science and Technology of China. He also worked at the University of Washington, Seattle, WA, USA, as a Visiting Scholar from August 2012 to August 2013. His research interests include speech processing, speech synthesis, voice conversion, and natural language processing. He was the recipient of the IEEE Signal Processing Society Young Author Best Paper Award in 2010. He is currently an Associate Editor of IEEE/ACM TRANSACTIONS ON AUDIO, SPEECH, AND LANGUAGE PROCESSING.
\end{IEEEbiography}

\begin{IEEEbiography}[{\includegraphics[width=1in,height=1.25in,clip,keepaspectratio]{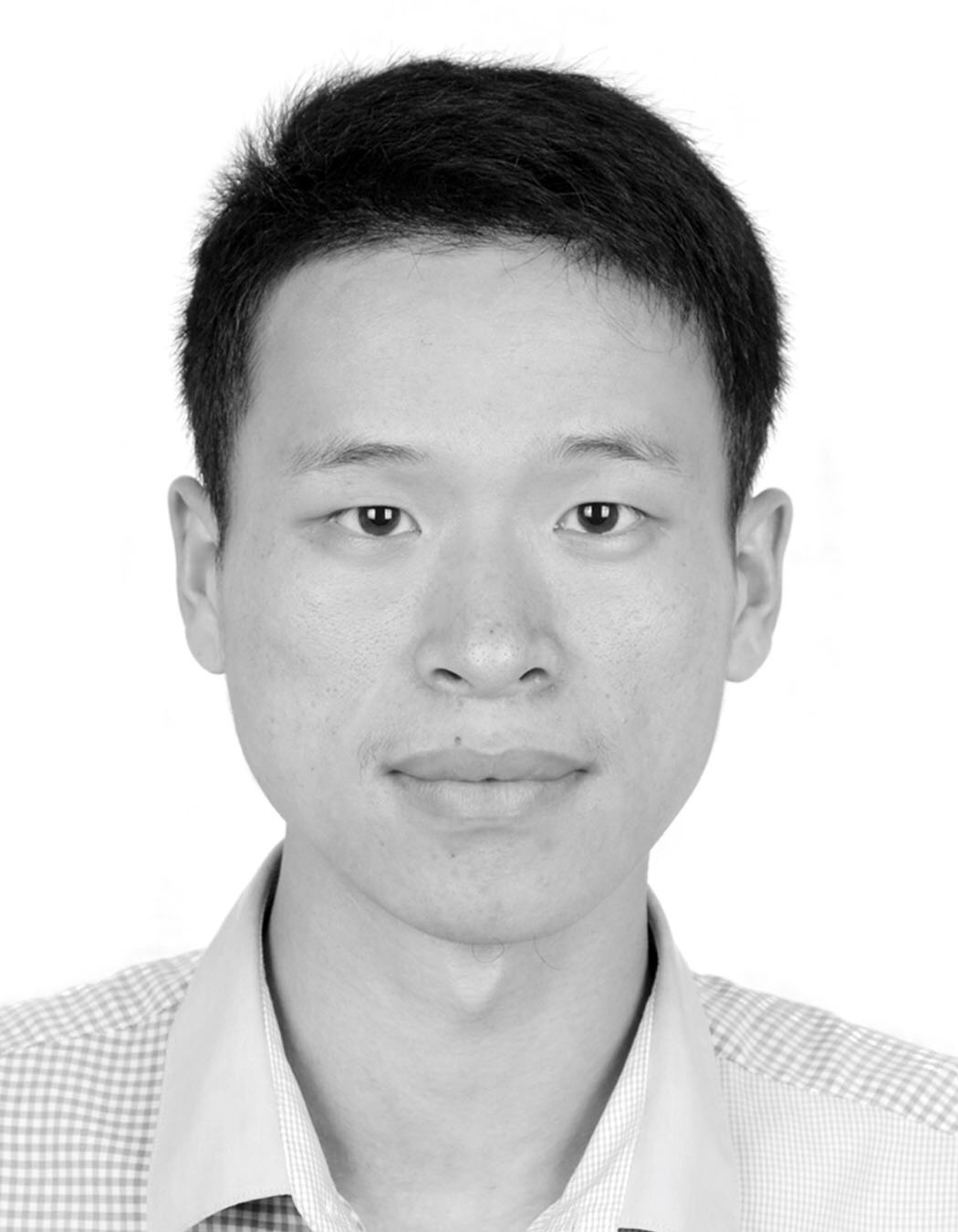}}]{Quan Liu}
  received a B.E. degree in electronic information engineering from Hohai University, Nanjing, China, in 2012, and a Ph.D. degree in signal and information processing from the University of Science and Technology of China, Hefei, China, in 2017.
  He is currently a postdoctoral researcher with University of Science and Technology of China, and iFLYTEK CO., LTD., Hefei, China.
  His research interests include machine learning for natural language processing.
\end{IEEEbiography}


\end{document}